\pdfoutput=1

\documentclass[11pt]{article}

\usepackage[final]{coling}
\usepackage{amsmath}

\usepackage[utf8]{inputenc} 
\usepackage[T1]{fontenc}    
\usepackage{hyperref}       
\usepackage{url}            
\usepackage{booktabs}       
\usepackage{amsfonts}       
\usepackage{nicefrac}       
\usepackage{microtype}      
\usepackage{xcolor}         
\usepackage{graphicx}
\usepackage{array}
\usepackage{multirow}
\newcolumntype{C}[1]{>{\centering\arraybackslash}m{#1}}
\definecolor{g}{HTML}{BFBFBF}

\usepackage{times}
\usepackage{latexsym}

\usepackage[T1]{fontenc}

\usepackage[utf8]{inputenc}

\usepackage{microtype}

\usepackage{inconsolata}

\usepackage{graphicx}

\title{Mix-of-Granularity: Optimize the Chunking Granularity for Retrieval-Augmented Generation}

\author{Zijie Zhong \\
  Shanghai AI Laboratory \\
  \texttt{zhongzijie@pjlab.org.cn} \\\And
  Hanwen Liu \\
  Beihang University \\
  \texttt{liuhwen@buaa.edu.cn} \\\And
  Xiaoya Cui \\
  Beihang University \\
  \texttt{xiaoya@buaa.edu.cn} \\\AND
  Xiaofan Zhang\footnotemark[1] \\
  Shanghai AI Laboratory \\
  \texttt{zhangxiaofan@pjlab.org.cn} \\\And
  Zengchang Qin\footnotemark[1] \\
  Beihang University, theSight Technology \\
  \texttt{zcqin@buaa.edu.cn} }

\begin{document}
\maketitle

\footnotetext[1]{Corresponding authors}

\begin{abstract}
Integrating information from various reference databases is a major challenge for Retrieval-Augmented Generation (RAG) systems because each knowledge source adopts a unique data structure and follows different conventions. Retrieving from multiple knowledge sources with one fixed strategy usually leads to under-exploitation of information. To mitigate this drawback, inspired by Mix-of-Expert, we introduce Mix-of-Granularity (MoG), a method that dynamically determines the optimal granularity of a knowledge source based on input queries using a router. The router is efficiently trained with a newly proposed loss function employing soft labels. We further extend MoG to MoG-Graph (MoGG), where reference documents are pre-processed as graphs, enabling the retrieval of distantly situated snippets. Experiments demonstrate that MoG and MoGG effectively predict optimal granularity levels, significantly enhancing the performance of the RAG system in downstream tasks. The code of both MoG and MoGG are released in \href{https://github.com/ZGChung/Mix-of-Granularity}{https://github.com/ZGChung/Mix-of-Granularity}.
\end{abstract}

\section{Introduction}

Retrieval-Augmented Generation (RAG) \cite{rag} has become a popular method for enhancing Large Language Models (LLMs). The core concept of RAG involves retrieving relevant information from external knowledge bases to provide additional context to the LLM, enabling it to generate more precise and grounded responses. RAG offers a promising and practical solution to mitigate LLMs' hallucinations because (1) it can be applied to any LLM, even those accessible only via APIs, and (2) the reference information is easy to modify or update. Many LLM-based products are supported by RAG systems, with examples spanning various industries such as customer service, advertising and marketing, education and e-learning, healthcare, and e-commerce and retailing \cite{ragproducts, medrag, ragtechdoc, rag4healthcare, passer}.

The quality of the retrieved snippets is crucial for the final generation, consequently, much research has focused on the retrieval phase. Currently, most RAG systems follow the Dual-Encoder Architecture \cite{dea} (DEA) paradigm, in which the reference documents are divided into small snippets (chunks), encoded by specific encoders, and then stored in the vector database (e.g. FAISS \cite{faiss} or Neo4j \cite{neo4j}) as embeddings. Thanks to its scalability, the DEA paradigm shows great potential for connecting LLMs with knowledge database of different formats, including knowledge graphs, textbooks, or Wikipedia articles.
\begin{figure*}[t]
  \centering
  \includegraphics[width=0.9\linewidth]{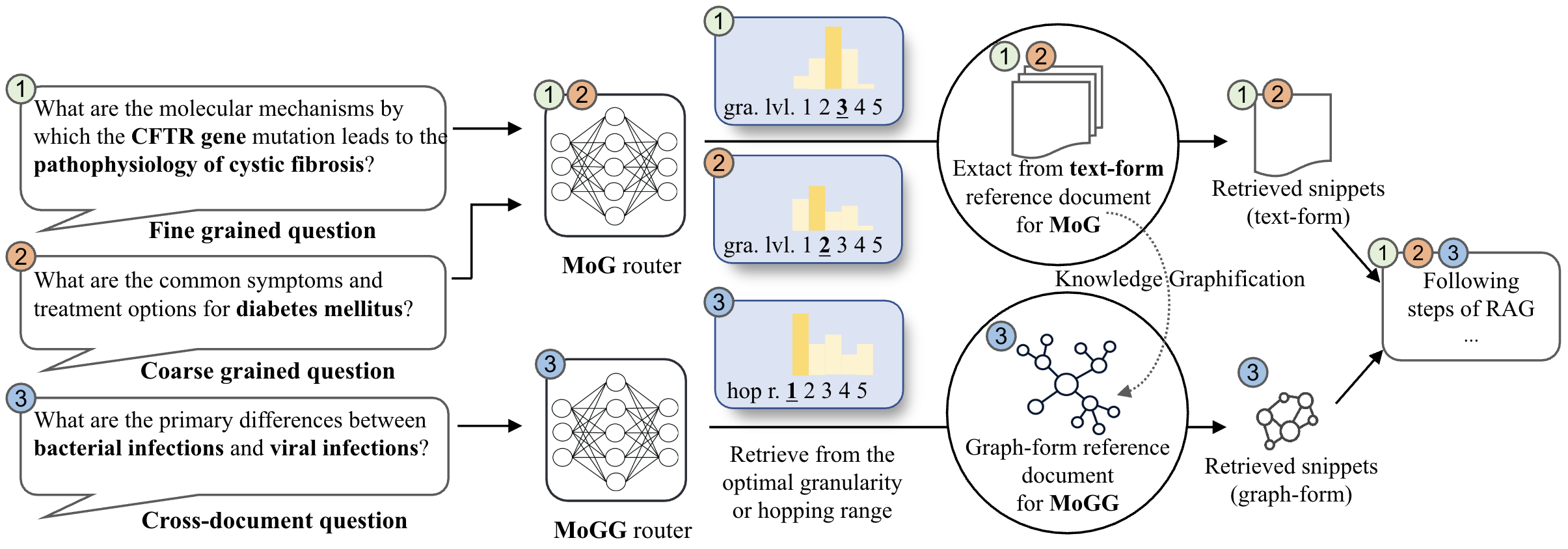}
  \caption{MoG automatically selects the optimal granularity when extracting information from the reference database (scenarios 1 and 2), achieving both high pertinence and coverage. When relevant information is dispersed across distant sections (scenario 3), the reference documents are pre-processed as graph, then MoGG is applied to retrieve these separate snippets from the best hopping range.}
  \label{fig:showcase}
\end{figure*}

Optimizing the chunk size of the reference knowledge database is essential for enhancing the precision and recall during the retrieval phase. However, this optimization presents significant challenges in the implementation of RAG systems for two primary reasons: (1) An optimal chunking size needs to be determined for each knowledge database due to their dissimilar data structures and information densities. For example, the optimal chunking size for medical textbooks is longer than the one for Medical Knowledge Graph Database (MKGD like Hetionet \cite{hetionet}), as Medical textbooks typically contain lengthy passages, while an MKGD consists of entities represented by shorter terms. (2) Even within a single knowledge database, using a uniform chunking size for all input queries can yield suboptimal retrieval results, as the queries themselves exhibit varying levels of granularity. As shown in Figure \ref{fig:showcase}), when the user asks about one disease (fine-grained question), chunking the reference document in finer granularity is better; whereas, when the user asks for broader information (coarse-grained question), a more coarse granularity is preferred. In practice, the optimal "uniform chunking size" is determined through parameter tuning, which is not only tedious but also does not guarantee high precision and recall.

Therefore, the community calls for a method to dynamically determine the optimal chunking size, for which we propose the Mix-of-Granularity (MoG). We draw inspiration from Mix-of-Experts \cite{moe}, which is a machine learning architecture that dynamically selects the most pertinent ``expert sub-network'' for each input token using a router module. Similarly, MoG involves leveraging a router to select the best granularity levels from which the reference snippets are extracted.

Even with this flexibility newly introduced, MoG still has difficulty dealing with broader queries that require distantly situated snippets, e.g. snippets stored in different knowledge databases. In such cases (cross-document question in Figure \ref{fig:showcase}), adjusting only the granularity is not helpful because the necessary information is so distantly located that it can never be covered by a reasonably large chunking window. To better answer these broader queries, we extend MoG to MoG-Graph (MoGG). In MoGG, the reference documents are pre-processed as a graph, allowing relevant snippets to be included as neighbors of each other, regardless of their distance in the original databases. This extension further improves the performance of cross-source retrieval.

When training MoG(G) under the supervised learning setting, the backward propagation is blocked by the top-$k$ selection at the end of the retrieval phase, which is a common practice in most RAG systems. To solve it, we introduce a loss function using soft labels. Soft labels are approximate training signals generated using offline algorithms or models like TF-IDF \cite{tfidf} or RoBERTa \cite{roberta}. With the soft labels, the top-$k$ selection is excluded from the training process, thus the issue of backward propagation is circumvented and the training is accelerated.

In conclusion, the main contributions of this work are: 

(1) We propose MoG, which dynamically determines the optimal granularity level for retrieval with the help of a router, achieving a balanced trade-off between precision and recall. 

(2) We extend MoG to MoGG by reorganizing the reference document in the form of a graph, thereby further improving the quality of cross-source retrieval involving multiple knowledge databases. 

(3) We introduce a loss function utilizing soft labels to overcome the challenges associated with training with top-$k$ selection.

\section{Related Work}
\subsection{Retrieval-Augmented Generation}
RAG \cite{rag} has emerged as a standard practice to enhance the LLMs by mitigating their problems of ``hallucinations'' and knowledge cut-off. A RAG system typically includes a Retriever that extracts relevant information from an external knowledge database, and a backbone LLM to generate grounded responses by in-context learning \cite{icl}. Previous retrieval-focused methods have evolved from retrieving tokens \cite{knnlm} or entities \cite{ease} to more complex structures like chunks \cite{ralm} or graphs \cite{surge}. Granularity matters a lot in retrieval, as coarse-granularity-retrieval yields more information with lower precision, while fine-granularity-retrieval offers comprehensive information at the cost of efficiency. More strategies like single \cite{skr, replug}, adaptive \cite{flare, raven}, or multiple retrieval \cite{atlas} are introduced to improve the retrieval phase's performance. Regarding the generation phase, various information fusion techniques are developed to integrate retrieval results to LLM in its input \cite{dsp}, intermediate \cite{retro}, or output layers \cite{intergen}.

\subsection{Chunking Optimization}
Optimal chunk size is crucial for RAG system as breaking down documents into small chunks is the first step to encode them. Naive chunking strategies, such as ``fixed-size'' chunking and ``recursive chunking,'' attempt to create snippets of identical size. Later works explored more chunking optimization techniques. One line of work focuses on increasing the recall of retrieval. For example, the ``Sliding Window Chunking'' \cite{slidingwindow} allows layered retrieval by merging globally related information across multiple processes. The ``Parent Document Retrieval'' \cite{pdr} retrieves using small chunks and returns larger blocks of context for later generation. Another line of work seeks to include more semantic information of the context to improve retrieval accuracy. ``Metadata Filtering'' \cite{metadatafiltering} leverages document metadata to filter snippets; ``Context-Enriched Chunking'' \cite{cec} breaks down information into segments and adds semantic summaries before retrieval; while ``Windowed Summarization Chunking'' \cite{cec} enriches each chunk with a windowed summary of the previous chunks. Many of these techniques, including MoG(G), are compatible with each other and can be combined to achieve better performance.

\subsection{Graph-Based Text Processing}
Graph-based text processing techniques combine research in graphs and text retrieval. Previous works exploit semantic similarities between small snippets (a sentence or several words) and reorganize the text material into a graph using Entity Recognition and Relation Construction algorithms \cite{text2kg, cyclegt, graphrnn}. Breaking the constraint of the single dimension of text corpus, these methods allow chunks of the same topic to be grouped as neighbors in a graph, thus show great potential in tasks requiring long context reasoning or multi-hop reasoning. For example, ``graph indexing'' \cite{ragsurvey} transforms entities and relationships into nodes and connections, improving the relevance of retrieved snippets significantly. RAPTOR \cite{raptor} organizes snippets as a tree (a special form of a graph) by recursively clustering them, where all non-leaf nodes correspond to summaries of their child nodes. This processing allows access to information at different granularity levels, which is beneficial to summarization tasks. In GMoE \cite{gmoe}, authors use different expert networks to handle hop-$1$, hop-$2$, and mixed hop-$1$ \& hop-$2$ neighbors of a node in a graph, which inspired our design of MoGG.

\section{Methodology}
\subsection{Preliminaries}
MoG(G) is designed to enhance the Retriever of a RAG system. A typical RAG system comprises a Retriever $\mathcal{R}$, a Generator $\mathcal{G}$, and a series of external reference corpus $\mathcal{K}=\{\mathcal{K}_1, \mathcal{K}_2, ..., \mathcal{K}_k\}$. A RAG system takes a user query $q$ as input, retrieves reference document pieces (snippets or chunks) from $\mathcal{K}$, and uses these snippets to help the $\mathcal{G}$ produce the final responses. A popular architecture for the Retriever is DEA \cite{dea}, where the query $q$ and all the snippets in $\mathcal{K}$ are encoded into embeddings ($e_q$ and $e_s$) using the same encoder $\mathcal{E}$. The extraction of relevant snippets is achieved by calculating the similarity between $e_q$ and $e_s$. The snippets with highest similarity scores are extracted and injected into the backbone LLM via prompt.

\subsection{Naive MoG}
\subsubsection{Multi-granularity Router}
\begin{figure*}[ht!]
  \centering
  \includegraphics[width=\linewidth]{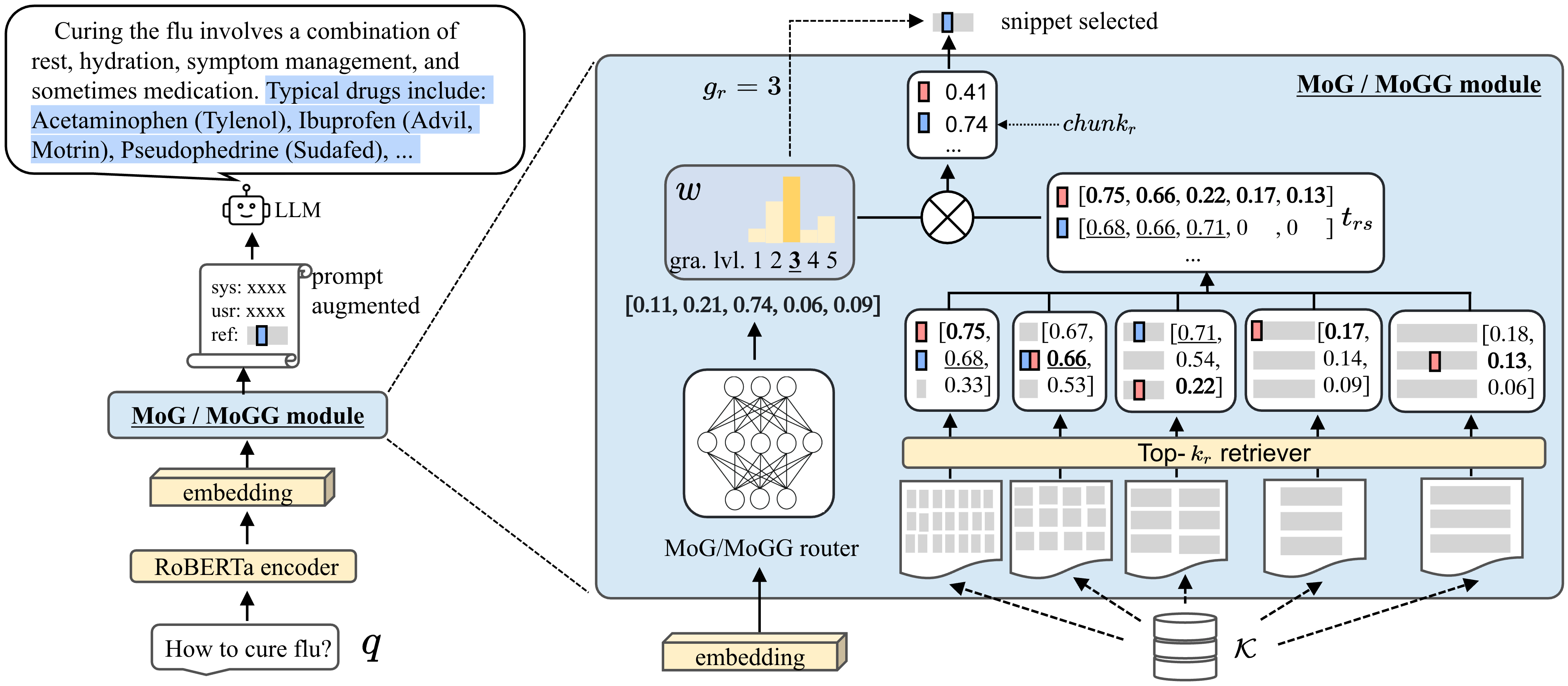}
  \caption{MoG mechanism prioritizes the chunks retrieved from optimal granularity level, which is determined by the router based on the user input query.}
  \label{fig:MoG_principal}
\end{figure*}

We apply the idea of Mix-of-Expert \cite{moe} (MoE) to automatically determine the best granularity level in the retrieval phase. In a MoE system, different input tokens are routed to the best expert network based on the weights output by the router. Similarly, in MoG, a router is trained to predict the importance weight of different granularity levels based on the user's input, so that the snippets from the best granularity level are prioritized. By employing such a routing optimization method, we can effectively adjust the chunk size according to different scenarios. Take one corpus $\mathcal{K}$ as example, its documents are chunked in $n_{gra}$ candidate granularities before the retrieval (in Figure \ref{fig:MoG_principal}, $n_{gra}$ = 5). Although some methods can dynamically adjust chunks using well-trained models, we prioritize the most commonly used and simplest chunking method from the perspective of practical efficiency. The snippets are chunked without overlap, and each chunk in $j$ ($j \in [2, n_{gra}]$) granularity level is formed with 2 adjacent chunks in $j-1$ granularity level (granularity level 1 is the finest). Each chunk is assigned a similarity score using BM25 \cite{bm25} with respect to user's input query $q$. In each granularity, $k_r$ most relevant snippets are extracted ($k_r$ = 3 in Figure \ref{fig:MoG_principal}), forming a pool of candidate snippets of size $n_{gra} \times k_r$. In parallel, $q$ is encoded via RoBERTa before being mapped to a weight $w$ (a vector of the same length as $n_{gra}$) by the router. The chunks' similarity scores are then weighted and summed with $w$. 

Empirically, directly retrieving the top snippets across different granularities based on their weighted similarity scores will give biased results, because the scores of more coarsed granularity levels are systematically higher (more discussions in Appendix \ref{sec:discussion_selection_process}). Therefore, we adopt another selection strategy involving two steps: (1) select the top-$k$ ($k$ = 1 in Figure \ref{fig:MoG_principal}) finest-grained chunk $chunk_{r}$; (2) retrieve the chunk containing $chunk_{r}$ from the optimal granularity level $g_r$. The selection process can be formalized as follows:

\begin{align}
    \label{eq: chunkr}
    chunk_{r} &= \underset{c \in \mathcal{C}}{\operatorname{top_k-argmax}} (t_{rs}(c) \cdot w); \\
    g_{r} &= \underset{g \in [1, n_{gra}] | w_{g} \neq 0}{\operatorname{argmax}} w_g\;
\end{align}

where $\mathcal{C}$ represents the set of all chunks of the finest granularity level, $t_{rs}(.)$ represents the relevance score of a chunk, and $w_g$ represents the $g$-th element in $w$.

Two snippets are highlighted in Figure \ref{fig:MoG_principal} as examples to help understanding. For each chunk of the finest granularity level, if it is retrieved among the top-$k_r$ snippets from one granularity level, its relevance score for that level is recorded as the actual relevance score (like the \textcolor{red}{red} chunk); otherwise, it is filled with 0 (like some zeroes padded for the \textcolor{blue}{blue} chunk). In this way, we get the relevance scores $t_{rs}$ of each finest snippet. The \textcolor{blue}{blue} chunk has highest weighted similarity score, it is therefore selected as $chunk_{r}$. In this example, the final snippet selected is the one extracted from 3rd granularity level containing $chunk_{r}$.

\subsubsection{Soft-labels}
In MoG, we train a Multi-Layer Perceptron (MLP) as the router in a supervised learning method. The input for the router is the embedding of the input query $q$ generated with RoBERTa \cite{roberta}, which is then mapped to $w$ by the router. The intrinsic training signal is the ``labels'' ($l$) in the Medical Question-Answering (MQA) datasets. In some MQA datasets, the ground truth snippets to be retrieved are provided, so we use them as ``labels'' directly; otherwise, the concatenation of the strings of the ``Question'' and ``Answer'' is used as the ``label''. A natural training objective is to maximize the semantic similarity between $chunk_r$ and $l$ by adjusting $w$. Such a label helps us to train the router to choose the best granularity based on the input query, which is a \textit{prior}, with solid \textit{posterior} information (the similarity between the ans and the text of different granularities). Unfortunately, this label can not be used to guide the training directly because there is a non-differentiable top-$k$ selection in the way. The soft labels are proposed to bypass the top-$k$ selection during the training: For each query $q$, the most relevant snippet ($S_{best}$) is retrieved from the reference documents of each granularity level with BM25 \cite{bm25}. The semantic similarity between each snippet in $S_{best}$ and the label $l$ is then calculated (with static models including TF-IDF \cite{tfidf}, RoBERTa \cite{roberta}, or hitrate score) and stored in $sim_{best}$. We create a soft label of 0.8 (resp. 0.2) for the most (resp. the second) similar snippet in $S_{best}$, and pad 0 for the other snippets. 

For example, the soft labels corresponding to $sim_{best,1}$ [0, 0.32, 0.11, 0.88, 0.45] and $sim_{best,2}$ [0.95, 0.07, 0.22, 0.11, 0.19] are $sl_1$ [0, 0, 0, 0.8, 0.2] and $sl_2$ [0.8, 0, 0.2, 0, 0], respectively. 

The values of the soft labels are designed to guide the router in distinguishing the relative importance among the granularity candidates. Empirically, setting these values to either [0.8, 0.2, 0] or [0.7, 0.3, 0] yields similar results. With the soft labels ($sl$) built, we can train the router to predict a high value (0.8) for the optimal granularity level, while conserving certain flexibility to choose the second-best granularity level. The router is trained by minimizing a Binary Cross Entropy loss function ($l_{bce}$):
\begin{equation}
l_{bce} = \sum_{i \in len(w)} -[ sl_i \cdot \log(w_i) + (1-sl_i) \cdot \log(1-w_i)].
\end{equation}

\subsection{MoGG: MoG with Graph-context}
With MoG, the adjacent snippets with relevant knowledge can be retrieved altogether by adjusting the granularity level. This method is particularly effective when information centered around the same topic is stored in adjoining sentences. However, in most cases, answering a complex question requires reasoning over information stored in different paragraphs or even different documents. A common solution is to perform more retrievals at a finer granularity level, retrieving only highly relevant small pieces to form a comprehensive reasoning chain. This approach is inconvenient because determining the optimal number of retrievals often involves manually adjusting $k$, which is challenging because the whole tuning process is time-consuming and $k$ can not go infinitely large.

\begin{figure*}[ht]
  \centering
  \includegraphics[width=0.8\linewidth]{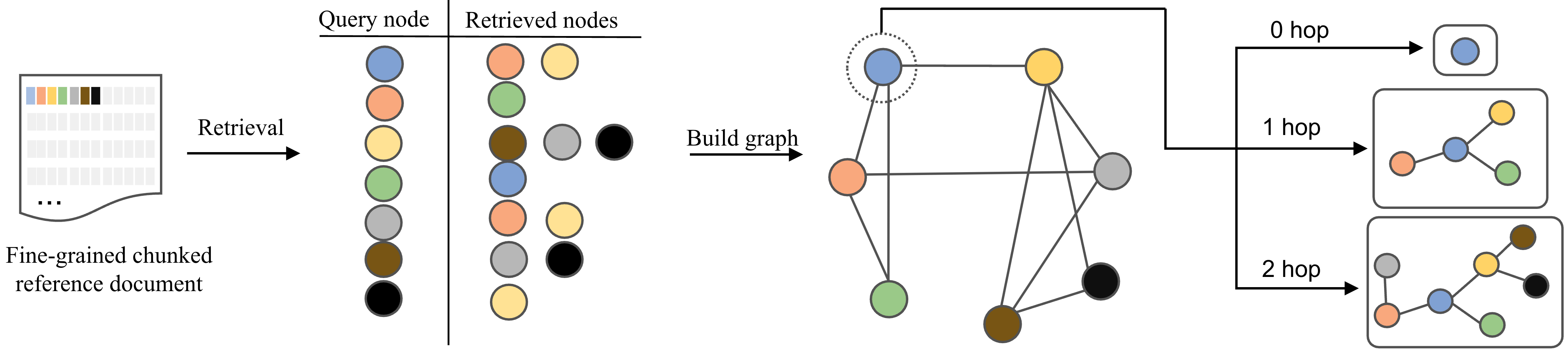}
  \caption{Pre-processing the reference document to form graphs. The concept of ``granularity level'' is changed into ``hopping range'' in graphs.}
  \label{fig:build_MoGG}
\end{figure*}

To overcome this challenge, an intuitive approach is to reorganize the reference document and group the relevant information together. 
Motivated by this, we propose a more applicable framework, MoGG, by enhancing MoG with a preprocessing step that organizes the documents in $\mathcal{K}$ as a graph. As illustrated in Figure \ref{fig:build_MoGG}, each document is initially split into small pieces consisting of one or two sentences, and each piece is treated as a separate node in the graph. To determine the edges in the graph, an index is first created with all these nodes. Then, each node is used as a ``query'' to search for the $k_{graph}$ (set as 3 in Figure \ref{fig:build_MoGG}) most relevant nodes using BM25. An edge is then added between two nodes if the similarity between them meets a predefined threshold $T_{graph}$.

This pre-processing extends the concept of ``context'' in linear text to ``neighbors'' in a graph. Also, granularity levels are adapted to ``hopping ranges'': in a non-graph setting, a larger granularity level corresponds to more adjacent sentences being grouped in a chunk; in a graph setting, a larger granularity level corresponds to nodes within a larger hopping range of a centered node being grouped as one chunk. To avoid context redundancy, duplicate nodes in the neighbors are considered only once. Similar to MoG, the documents in the external knowledge database $\mathcal{K}$ are chunked with $n_{gra}$ different hopping ranges and then encoded into embeddings, while the rest of the MoG remains unchanged. By transforming the documents into a graph and defining granularity based on hopping ranges, MoGG effectively captures dispersed relevant information, allowing for more comprehensive and efficient retrieval.

\section{Experiments}
\subsection{Corpus and Medical QA datasets}
A reference knowledge database used in a RAG system is often termed a ``corpus''. To form the corpora, data from various sources were collected, including the widely-used PubMed \cite{pubmed} corpus for all biomedical abstracts, the StatPearls \cite{statpearls} corpus for clinical decision support, the Textbooks\cite{textbooks} corpus covering medical textbook knowledge, and the Wikipedia \cite{wikipedia_data} corpus for general knowledge. The four corpora are combined to form a larger corpus, named ``MedCorp''. For each corpus used in our experiments, there are 5 granularity levels tested. The chunking size of the second granularity level is set to be the same as the one in MedRAG to facilitate the comparison of the results. The chunking size of the \{1st, 3rd, 4th, 5th\} granularity level is set to be \{1/2, 2, 4, 8\} times of the one in MedRAG.

The performance of the RAG system are evaluated using five MQA datasets following MIRAGE benchmark \cite{medrag}: MMLU-Med \cite{mmlu}, MedQA-US \cite{medqa}, MedMCQA \cite{medmcqa}, PubMedQA* \cite{pubmedqa}, and BioASQ-Y/N \cite{bioasq}. Specifically, only the biomedical questions are kept and the ground truth supporting contexts are removed during testing. 

The detailed descriptions of each copora and each QA dataset, as well as their important statistics, are included in Appendix \ref{sec:qa_datasets}. We have begun to test our method using medical question-answering datasets, as they represent a knowledge-intensive domain. We posit that significant improvements demonstrated by the tests on this knowledge-intensive field suggest MoG's potential effectiveness in other domains with lower knowledge dependencies and higher error tolerances.

\subsection{Experiment Setup}
All backbone LLMs, whether accessed via API or local deployment, are run under off-the-shelf settings. The exact versions of the backbone LLMs are listed in Appendix \ref{sec:llm_versions}. Experiments are conducted on Nvidia GeForce 3090 and 4090 GPUs. The code is written using the PyTorch framework, utilizing an Adam optimizer with a learning rate of 0.001. Each training job is run until the convergence of the loss value. During the experiment, there are two top-k selections. Unless specified, when retrieving snippets from each corpus, we select the top-3 snippets; when all the snippets are retrieved from each corpus, we select the top-2 snippets with the highest relevance scores to pass to the backbone LLM. When snippets are too long, they are truncated automatically to fit the LLM's context window size. The router requires approximately 12GB of GPU memory for training and 6GB for inference. Utilizing a caching mechanism, we efficiently completed 35 training sessions and over a hundred inferences, each training session taking around 4 hours for 1000 epochs.

\subsection{Performance of MoG on MQA Task}
As mentioned above, we test the effectiveness of MoG on MQA datasets. For each question, the RAG system is tasked with choosing the best answer(s) from the given options. The performance of the entire RAG system is measured by the Exact Matching accuracy of the answers. To prevent knowledge leakage, only the question is used (options excluded) to retrieve reference documents from the external knowledge database. The router of MoG guides the retrieval system to choose the optimal granularity. Qualitatively speaking, when tested on different datasets, the router trained with Textbook corpus shows a preference for different granularity levels. For instance, on the PubMedQA dataset, the finest granularity snippets are selected most frequently. This is because the questions in PubMedQA are typically precise and can be answered with short reference snippets. An example result figure is shown in Figure \ref{fig:granularity distribution}, with a more detailed discussion included in Appendix \ref{sec:qualitative_res}.

\begin{figure*}[h!]
  \centering
  \includegraphics[width=0.8\linewidth]{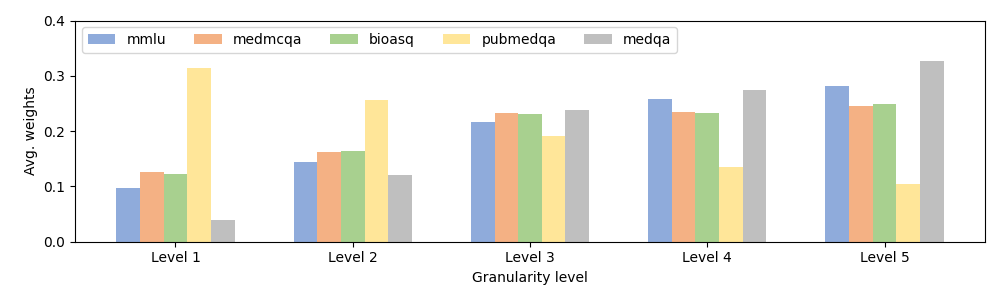}
  \caption{Averaged weights of different granularity levels on different QA datasets}
  \label{fig:granularity distribution}
\end{figure*}

\begin{table*}[ht]
  \centering
  \small
  \begin{tabular}{C{1cm}C{1.1cm}C{1.6cm}C{1.6cm}C{1.6cm}C{1.6cm}C{1.6cm}C{0.8cm}} 
  \toprule
\multirow{2}{*}{LLM}       & \multirow{2}{*}{Method} & \multicolumn{6}{c}{MIRAGE Benchmark Dataset (Acc.)} \\

\cmidrule(r){3-8}
                           &  & \small{MMLU}  & \small{MedQA} & \small{MedMCQA}   & \small{PubMedQA}  & \small{BioASQ}    & \small{Avg.} \\
   \midrule
\multirow{3}{*}{GLM3}       & CoT                    &0.4356\textcolor{g}{$\pm$0.04}         &0.3451\textcolor{g}{$\pm$0.03}         &0.3250\textcolor{g}{$\pm$0.02}
                                                     &0.3400\textcolor{g}{$\pm$0.05}         &0.5081\textcolor{g}{$\pm$0.04}         &0.3908                        \\
                            & MedRAG                 &0.4950\textcolor{g}{$\pm$0.04}         &\textbf{0.3804}\textcolor{g}{$\pm$0.03}&0.3632\textcolor{g}{$\pm$0.02}
                                                     &0.5100\textcolor{g}{$\pm$0.05}         &0.6532\textcolor{g}{$\pm$0.04}         &0.4804                         \\
                            & MoG                    &\textbf{0.5198}\textcolor{g}{$\pm$0.04}&0.3529\textcolor{g}{$\pm$0.03}         &\textbf{0.3716}\textcolor{g}{$\pm$0.02}
                                                     &\textbf{0.5400}\textcolor{g}{$\pm$0.05}&\textbf{0.6774}\textcolor{g}{$\pm$0.04}&\textbf{0.4923}                \\
                        
   \midrule
\multirow{3}{*}{GPT-3.5}  & CoT                      &\textbf{0.7525}\textcolor{g}{$\pm$0.03}&0.6118\textcolor{g}{$\pm$0.03}&\textbf{0.5890}\textcolor{g}{$\pm$0.02}&0.5200\textcolor{g}{$\pm$0.05}&0.7339\textcolor{g}{$\pm$0.04}&0.6571      \\
                            & MedRAG                 &0.6931\textcolor{g}{$\pm$0.03}&\textbf{0.6510}\textcolor{g}{$\pm$0.03}&0.4671\textcolor{g}{$\pm$0.02}&0.6000\textcolor{g}{$\pm$0.05}&\textbf{0.8306}\textcolor{g}{$\pm$0.03}&0.6484      \\
                           & MoG                     &0.7129\textcolor{g}{$\pm$0.03}&0.6471\textcolor{g}{$\pm$0.03}&0.5532\textcolor{g}{$\pm$0.02}&\textbf{0.6200}\textcolor{g}{$\pm$0.05}&0.7823\textcolor{g}{$\pm$0.04}&\textbf{0.6631}      \\
                          
   \midrule
\multirow{3}{*}{InternLM}    & CoT                   &\textbf{0.7426}\textcolor{g}{$\pm$0.03}&\textbf{0.6118}\textcolor{g}{$\pm$0.03}&0.5269\textcolor{g}{$\pm$0.02}&0.3500\textcolor{g}{$\pm$0.05}&0.7258\textcolor{g}{$\pm$0.04}&0.5914      \\
                            & MedRAG                 &0.6040\textcolor{g}{$\pm$0.04}&0.5294\textcolor{g}{$\pm$0.03}&0.3847\textcolor{g}{$\pm$0.03}&0.4300\textcolor{g}{$\pm$0.05}&\textbf{0.7661}\textcolor{g}{$\pm$0.04}&0.5428      \\
                           & MoG                     &0.7129\textcolor{g}{$\pm$0.03}    &0.5961\textcolor{g}{$\pm$0.03}     &\textbf{0.5436}\textcolor{g}{$\pm$0.02}        &\textbf{0.4100}\textcolor{g}{$\pm$0.05}        
                                                     &\textbf{0.7661}\textcolor{g}{$\pm$0.04}   &\textbf{0.6057}     \\
                        
  \midrule
\multirow{3}{*}{Llama3}      & CoT                   &0.7079\textcolor{g}{$\pm$0.03}&\textbf{0.6431}\textcolor{g}{$\pm$0.03}&\textbf{0.5663}\textcolor{g}{$\pm$0.02}&0.5500\textcolor{g}{$\pm$0.05}&0.7258\textcolor{g}{$\pm$0.04}&0.6386      \\
                            & MedRAG                 &0.6040\textcolor{g}{$\pm$0.03}&0.5725\textcolor{g}{$\pm$0.03}&0.4313\textcolor{g}{$\pm$0.02}&0.5600\textcolor{g}{$\pm$0.05}&0.7823\textcolor{g}{$\pm$0.04}&0.5900      \\
                           & MoG                     &\textbf{0.7228}\textcolor{g}{$\pm$0.03}&0.6000\textcolor{g}{$\pm$0.03}&0.5627\textcolor{g}{$\pm$0.02}&\textbf{0.6400}\textcolor{g}{$\pm$0.05}&\textbf{0.7984}\textcolor{g}{$\pm$0.04}&\textbf{0.6648}      \\
                       
   \midrule
\multirow{3}{*}{Qwen1.5}      & CoT                  &0.4604\textcolor{g}{$\pm$0.04}&0.3255\textcolor{g}{$\pm$0.03}&0.3883\textcolor{g}{$\pm$0.02}&0.2000\textcolor{g}{$\pm$0.03}&0.5484\textcolor{g}{$\pm$0.04}&0.3845      \\
                            & MedRAG                 &0.5594\textcolor{g}{$\pm$0.03}&\textbf{0.4353}\textcolor{g}{$\pm$0.03}&0.4038\textcolor{g}{$\pm$0.02}&0.3400\textcolor{g}{$\pm$0.05}&0.5403\textcolor{g}{$\pm$0.04}&0.4558      \\
                           & MoG                     &\textbf{0.5941}\textcolor{g}{$\pm$0.04}&0.4235\textcolor{g}{$\pm$0.03}&\textbf{0.4301}\textcolor{g}{$\pm$0.02}&\textbf{0.4700}\textcolor{g}{$\pm$0.05}&\textbf{0.6694}\textcolor{g}{$\pm$0.04}&\textbf{0.5174}      \\
                         
   \bottomrule
\end{tabular}
  \caption{Accuracy of Medical Question-Answering task with MoG (trained with MedCorp), best results in \textbf{bold}.}
  \label{tab:mog_result_table}
\end{table*}

MoG is integrated into one same RAG system, with which we conduct the MQA task. Backbone LLMs are altered to cover some of the popular ones, such as ChatGPT \cite{gpt3}, InternLM2 \cite{internlm2}, Llama3 \cite{llama3}, GLM3 \cite{glm}, and Qwen1.5 \cite{qwen}. The router is trained with the soft labels built using RoBERTa \cite{roberta}, this choice is justified by the experiment in Appendix \ref{sec:soft_labels}. To investigate the effect of varying the number of candidate snippets on RAG system performance, we conducted experiments with different snippet counts (details are provided in Appendix \ref{sec:rag_k}). The retriever is fixed as BM25 \cite{bm25}, with a further discussion on the performance of different retrievers included in Appendix \ref{sec:choice_retriever}. In Table \ref{tab:mog_result_table} we present the results obtained with the router trained with MedCorp corpus (the results obtained with routers trained on four single corpora are presented in Appendix \ref{sec:mog_other_results}).The results are compared with two baselines: CoT and MedRAG. CoT baseline adopts Chain-of-Thought \cite{cot} prompting and does not leverage any external knowledge database to enhance its response. MedRAG baseline is a simple RAG system with only 1 candidate granularity level introduced in MedRAG paper \cite{medrag}.

The results demonstrate that MoG consistently enhanced the performance of the RAG system across different backbone models when compared with MedRAG, though not necessarily better than CoT. The reason is that the RAG system we used has no noise filters or any quality control mechanism, thus the noise is injected along with the knowledge via prompts. A detailed analysis of the number of samples improved or degraded by the application of MoG is included in the Appendix \ref{sec:analysis_sample}, in which we manually verified that the majority of degradation is caused by noise. We also find that MoG improves the accuracy score more when applied on weaker LLMs (like ChatGLM and Qwen), probably because they have less knowledge stored in their internal parameters and, thus could benefit more from the retrieved snippets.

\begin{table*}[h]
  \centering
  \small
  \begin{tabular}{C{1cm}C{1cm}C{1.6cm}C{1.6cm}C{1.6cm}C{1.6cm}C{1.6cm}C{0.8cm}}
  \toprule
\multirow{2}{*}{LLM}       & \multirow{2}{*}{Method} & \multicolumn{6}{c}{MIRAGE Benchmark Dataset (Acc.)} \\

\cmidrule(r){3-8}
                           &                        & \small{MMLU}  & \small{MedQA} & \small{MedMCQA}   & \small{PubMedQA}  & \small{BioASQ}    & \small{Avg.} \\
   \midrule
\multirow{3}{*}{GLM3}       & CoT                    &0.4356\textcolor{g}{$\pm$0.04}&0.3451\textcolor{g}{$\pm$0.03}&0.3250\textcolor{g}{$\pm$0.02}&0.3400\textcolor{g}{$\pm$0.05}&0.5081\textcolor{g}{$\pm$0.04}&0.3908     \\
                           & MedRAG                  &0.4802\textcolor{g}{$\pm$0.04}         &\textbf{0.3569}\textcolor{g}{$\pm$0.03} &\textbf{0.3811}\textcolor{g}{$\pm$0.02} &0.3600\textcolor{g}{$\pm$0.05}          &0.5565\textcolor{g}{$\pm$0.04}          &0.4269      \\
                           & MoG                     &\textbf{0.5545}\textcolor{g}{$\pm$0.04}&0.2941\textcolor{g}{$\pm$0.03}          &0.3548\textcolor{g}{$\pm$0.02}          &\textbf{0.4700}\textcolor{g}{$\pm$0.05} &\textbf{0.5726}\textcolor{g}{$\pm$0.04} &\textbf{0.4492}      \\
                           & MoGG                    &0.5347\textcolor{g}{$\pm$0.04}         &0.3176\textcolor{g}{$\pm$0.03}          &0.3608\textcolor{g}{$\pm$0.02}          &0.4500\textcolor{g}{$\pm$0.05}          &0.5645\textcolor{g}{$\pm$0.04}          &0.4455      \\       
   \midrule
\multirow{3}{*}{GPT-3.5}  & CoT                      &0.7525\textcolor{g}{$\pm$0.03}&0.6118\textcolor{g}{$\pm$0.03}&\textbf{0.5890}\textcolor{g}{$\pm$0.02}&\textbf{0.5200}\textcolor{g}{$\pm$0.05}&\textbf{0.7339}\textcolor{g}{$\pm$0.04}&\textbf{0.6571}      \\
                           & MedRAG                  &0.7277\textcolor{g}{$\pm$0.03}         &0.6745\textcolor{g}{$\pm$0.03}          &0.4468\textcolor{g}{$\pm$0.02}          &0.2600\textcolor{g}{$\pm$0.04}          &0.5161\textcolor{g}{$\pm$0.04}          &0.5250      \\
                           & MoG                     &0.7525\textcolor{g}{$\pm$0.03}         &0.6667\textcolor{g}{$\pm$0.03}          &0.5603\textcolor{g}{$\pm$0.02}          &\textbf{0.5200}\textcolor{g}{$\pm$0.05} &0.7016\textcolor{g}{$\pm$0.04}          &0.6122      \\
                           & MoGG                    &\textbf{0.7673}\textcolor{g}{$\pm$0.03}&\textbf{0.6784}\textcolor{g}{$\pm$0.03}          &0.5233\textcolor{g}{$\pm$0.02}          &0.4700\textcolor{g}{$\pm$0.05}          &0.6452\textcolor{g}{$\pm$0.04}          &0.6168      \\
   \midrule
\multirow{3}{*}{InternLM}    & CoT                   &\textbf{0.7426}\textcolor{g}{$\pm$0.03}&\textbf{0.6118}\textcolor{g}{$\pm$0.03}&0.5269\textcolor{g}{$\pm$0.02}&\textbf{0.3500}\textcolor{g}{$\pm$0.05}&0.7258\textcolor{g}{$\pm$0.04}&\textbf{0.5914}      \\
                           & MedRAG                  &0.6188\textcolor{g}{$\pm$0.03}         &0.5569\textcolor{g}{$\pm$0.03}          &0.3118\textcolor{g}{$\pm$0.02}          &0.1400\textcolor{g}{$\pm$0.03}          &0.4839\textcolor{g}{$\pm$0.04}          &0.4223      \\
                           & MoG                     &0.7277\textcolor{g}{$\pm$0.03}         &0.5725\textcolor{g}{$\pm$0.03}          &\textbf{0.5281}\textcolor{g}{$\pm$0.02} &0.3400\textcolor{g}{$\pm$0.05}          &\textbf{0.7339}\textcolor{g}{$\pm$0.04} &0.5804      \\
                           & MoGG                    &0.7228\textcolor{g}{$\pm$0.03}         &0.5882\textcolor{g}{$\pm$0.03}          &0.5173\textcolor{g}{$\pm$0.02}          &0.3400\textcolor{g}{$\pm$0.05}          &\textbf{0.7339}\textcolor{g}{$\pm$0.04} &0.5804      \\
  \midrule
\multirow{3}{*}{Llama3}      & CoT                   &0.7079\textcolor{g}{$\pm$0.03}&\textbf{0.6431}\textcolor{g}{$\pm$0.03}&\textbf{0.5663}\textcolor{g}{$\pm$0.02}&\textbf{0.5500}\textcolor{g}{$\pm$0.05}&0.7258\textcolor{g}{$\pm$0.04}&\textbf{0.6386}      \\
                           & MedRAG                  &0.6485\textcolor{g}{$\pm$0.03}         &0.5961\textcolor{g}{$\pm$0.03}          &0.4146\textcolor{g}{$\pm$0.02}          &0.3800\textcolor{g}{$\pm$0.05}          &0.5242\textcolor{g}{$\pm$0.04}          &0.5127      \\
                           & MoG                     &\textbf{0.7228}\textcolor{g}{$\pm$0.03}&0.6196\textcolor{g}{$\pm$0.03}          &0.5484\textcolor{g}{$\pm$0.02}          &0.5100\textcolor{g}{$\pm$0.05}          &0.7097\textcolor{g}{$\pm$0.04}          &0.6221      \\
                           & MoGG                    &0.7030\textcolor{g}{$\pm$0.03}         &0.5961\textcolor{g}{$\pm$0.03}          &0.5460\textcolor{g}{$\pm$0.02}          &0.5200\textcolor{g}{$\pm$0.05}          &\textbf{0.7661}\textcolor{g}{$\pm$0.04} &0.6262      \\
   \midrule
\multirow{3}{*}{Qwen1.5}      & CoT                  &0.4604\textcolor{g}{$\pm$0.04}&0.3255\textcolor{g}{$\pm$0.03}&0.3883\textcolor{g}{$\pm$0.02}&0.2000\textcolor{g}{$\pm$0.03}&0.5484\textcolor{g}{$\pm$0.04}&0.3845      \\
                           & MedRAG                  &\textbf{0.5941}\textcolor{g}{$\pm$0.03}&0.4000\textcolor{g}{$\pm$0.03}          &0.3835\textcolor{g}{$\pm$0.02}          &\textbf{0.3300}\textcolor{g}{$\pm$0.05} &0.4919\textcolor{g}{$\pm$0.04}          &0.4399      \\
                           & MoG                     &0.5792\textcolor{g}{$\pm$0.03}         &0.3843\textcolor{g}{$\pm$0.03}          &0.4110\textcolor{g}{$\pm$0.02}          &\textbf{0.3300}\textcolor{g}{$\pm$0.05} &0.6129\textcolor{g}{$\pm$0.04}          &0.4635      \\
                           & MoGG                    &0.5594\textcolor{g}{$\pm$0.03}         &\textbf{0.4314}\textcolor{g}{$\pm$0.03} &\textbf{0.4480}\textcolor{g}{$\pm$0.02} &0.3000\textcolor{g}{$\pm$0.05}          &\textbf{0.6371}\textcolor{g}{$\pm$0.04} &\textbf{0.4752}      \\
   \bottomrule
\end{tabular}
  \caption{Accuracy of Medical Question-Answering task with MoGG (trained with Textbooks), best results in \textbf{bold}.}
  \label{tab:mogg_result_table_textbooks}
\end{table*}

\subsection{Performance of MoGG on MQA Task}
Similarly to the experiment of MoG, we test the performance of MoGG on MQA datasets. We tested only the performance of MoGG with routers trained on Textbooks and StatPearls corpora because training with the other two much larger corpora is too time-consuming. In Table \ref{tab:mogg_result_table_textbooks} we present the results obtained when trained with Textbook corpus (results obtained with StatPearls showed in Appendix \ref{sec:mogg_statpearls} with similar patterns). From the table, we can tell that MoGG can further improve the averaged accuracy scores. By comparing Table \ref{tab:mog_result_table} and Table \ref{tab:mogg_result_table_textbooks}, we can find that, even when trained with significantly fewer samples (Textbooks corpus is a tiny subset of MedCorp corpus, accounting for only about 0.2\% of all the snippets in MedCorp), MoGG brings more significant improvement in terms of the averaged accuracy score with respect to MedRAG than MoG. This finding highlights that MoGG is more efficient than MoG thanks to its flexible way of organizing the reference snippets (in the form of a graph). We conducted an ablation test with Llama 3 and the results are included in Appendix \ref{sec:mogg_ablation}.

There is a general performance drop in the metrics in Table \ref{tab:mogg_result_table_textbooks} compared to Tabel \ref{tab:mog_result_table} because the Textbooks corpus is only a tiny subset of the MedCorp used in Table \ref{tab:mog_result_table}). To facilitate the comparison, we conducted the same experiment in Table \ref{tab:mog_result_table} with only the Textbooks corpus, and the detailed results are reported in Appendix \ref{sec:mog_only_textbooks}.

\subsection{Execution time and storage efficiency}
MoG(G) is proposed to increase the precision and recall of the retrieval phase at a reasonable cost of computational efficiency because the quality of the retrieved chunks is prioritized for application scenarios like the medical environment. We measured the average inference time with the different number of candidate granularity levels, and the results show that increasing the number of granularity levels will only increase a marginal increase in execution time. (Details in Appendix \ref{sec:time_consumption})

In terms of storage, while additional space is required to store the embeddings of the corpus at different granularities, we only need to store one copy of the corpus and five sets of embeddings. This engineering optimization results in a space requirement of only 2.7 times the size of the original corpus. We believe this represents an acceptable overhead. Furthermore, in our RAG system specialized in the medical domain, a corpus containing about 10GB of plain text can already cover a wide range of questions. Thus, we believe this will not be an obstacle to MoG(G)'s wider application.

We believe this overhead of computing resources added is worthwhile because in most cases, MoG performs much better than MedRAG and CoT baselines. On smaller models, MoG shows an average improvement of 5\% compared to MedRAG and 8.7\% compared to CoT. Given the dataset size of approximately 7000, these improvements are statistically and practically significant.

\section{Limitations and Broader Impacts}
Our work serves as an early trial in dynamic chunking strategy, with the following major directions for improvement. (1) MoG(G)'s candidate granularity levels are manually assigned. It could be more efficient if an algorithm automatically set these granularity levels to avoid excessive grid-searching for parameter optimization. (2) Current router uses only the semantic information of the input query to predict the best granularity level. (3) Previous studies have demonstrated that the use of length normalization is crucial in information retrieval-related fields. This paper primarily focuses on applying the concept of Mix-of-Experts to this application scenario. In future work, we will also take related issues into consideration. Incorporating more information (like query type or expected response length) into the router can potentially improve the results. 
However, the router also introduces a new security risk: a compromised router could redirect knowledge retrieval to malicious sources, injecting incorrect or even harmful information into the backbone LLM. Therefore, it is crucial to protect and monitor the router to mitigate this risk.
    
\section{Conclusion}
In this work, we present MoG, a mechanism to dynamically choose the best granularity when retrieving information from an external knowledge database. When applied to a RAG system, MoG helps retrieve more relevant information while reducing noise. MoG is further extended as MoGG, where reference documents are pre-processed as graphs. This extension allows distantly situated information to be retrieved simultaneously, overcoming the limitations of a fixed top-\( k \) selection strategy. Finally, we introduce a soft label guided loss function to address the difficulty of backward propagation with top-\( k \) selection, which could benefit future research.

\section*{Acknowledgements}
We would like to express our gratitude to all the supervisors and colleagues at Shanghai Artificial Intelligence Laboratory for their invaluable insights, feedback, and support throughout the research process. We also thank the reviewers for their constructive comments, which greatly improved the quality of this paper.




\clearpage
\bibliography{custom}

\begin{thebibliography}{56}
\providecommand{\natexlab}[1]{#1}

\bibitem[{AI(2024)}]{llama3}
Meta AI. 2024.
\newblock Introducing meta llama 3: The most capable openly available llm to date.
\newblock \emph{Meta document}.

\bibitem[{Bai et~al.(2023)Bai, Bai, Chu, Cui, Dang, Deng, Fan, Ge, Han, Huang, Hui, Ji, Li, Lin, Lin, Liu, Liu, Lu, Lu, Ma, Men, Ren, Ren, Tan, Tan, Tu, Wang, Wang, Wang, Wu, Xu, Xu, Yang, Yang, Yang, Yang, Yao, Yu, Yuan, Yuan, Zhang, Zhang, Zhang, Zhang, Zhou, Zhou, Zhou, and Zhu}]{qwen}
Jinze Bai, Shuai Bai, Yunfei Chu, Zeyu Cui, Kai Dang, Xiaodong Deng, Yang Fan, Wenbin Ge, Yu~Han, Fei Huang, Binyuan Hui, Luo Ji, Mei Li, Junyang Lin, Runji Lin, Dayiheng Liu, Gao Liu, Chengqiang Lu, Keming Lu, Jianxin Ma, Rui Men, Xingzhang Ren, Xuancheng Ren, Chuanqi Tan, Sinan Tan, Jianhong Tu, Peng Wang, Shijie Wang, Wei Wang, Shengguang Wu, Benfeng Xu, Jin Xu, An~Yang, Hao Yang, Jian Yang, Shusheng Yang, Yang Yao, Bowen Yu, Hongyi Yuan, Zheng Yuan, Jianwei Zhang, Xingxuan Zhang, Yichang Zhang, Zhenru Zhang, Chang Zhou, Jingren Zhou, Xiaohuan Zhou, and Tianhang Zhu. 2023.
\newblock \href {https://arxiv.org/abs/2309.16609} {Qwen technical report}.
\newblock \emph{Preprint}, arXiv:2309.16609.

\bibitem[{Borgeaud et~al.(2022)Borgeaud, Mensch, Hoffmann, Cai, Rutherford, Millican, van~den Driessche, Lespiau, Damoc, Clark, de~Las~Casas, Guy, Menick, Ring, Hennigan, Huang, Maggiore, Jones, Cassirer, Brock, Paganini, Irving, Vinyals, Osindero, Simonyan, Rae, Elsen, and Sifre}]{retro}
Sebastian Borgeaud, Arthur Mensch, Jordan Hoffmann, Trevor Cai, Eliza Rutherford, Katie Millican, George van~den Driessche, Jean-Baptiste Lespiau, Bogdan Damoc, Aidan Clark, Diego de~Las~Casas, Aurelia Guy, Jacob Menick, Roman Ring, Tom Hennigan, Saffron Huang, Loren Maggiore, Chris Jones, Albin Cassirer, Andy Brock, Michela Paganini, Geoffrey Irving, Oriol Vinyals, Simon Osindero, Karen Simonyan, Jack~W. Rae, Erich Elsen, and Laurent Sifre. 2022.
\newblock \href {https://arxiv.org/abs/2112.04426} {Improving language models by retrieving from trillions of tokens}.
\newblock \emph{Preprint}, arXiv:2112.04426.

\bibitem[{Brown et~al.(2020)Brown, Mann, Ryder, Subbiah, Kaplan, Dhariwal, Neelakantan, Shyam, Sastry, Askell, Agarwal, Herbert-Voss, Krueger, Henighan, Child, Ramesh, Ziegler, Wu, Winter, Hesse, Chen, Sigler, Litwin, Gray, Chess, Clark, Berner, McCandlish, Radford, Sutskever, and Amodei}]{gpt3}
Tom Brown, Benjamin Mann, Nick Ryder, Melanie Subbiah, Jared~D Kaplan, Prafulla Dhariwal, Arvind Neelakantan, Pranav Shyam, Girish Sastry, Amanda Askell, Sandhini Agarwal, Ariel Herbert-Voss, Gretchen Krueger, Tom Henighan, Rewon Child, Aditya Ramesh, Daniel Ziegler, Jeffrey Wu, Clemens Winter, Chris Hesse, Mark Chen, Eric Sigler, Mateusz Litwin, Scott Gray, Benjamin Chess, Jack Clark, Christopher Berner, Sam McCandlish, Alec Radford, Ilya Sutskever, and Dario Amodei. 2020.
\newblock \href {https://proceedings.neurips.cc/paper_files/paper/2020/file/1457c0d6bfcb4967418bfb8ac142f64a-Paper.pdf} {Language models are few-shot learners}.
\newblock In \emph{Advances in Neural Information Processing Systems}, volume~33, pages 1877--1901. Curran Associates, Inc.

\bibitem[{Cai et~al.(2024)Cai, Cao, Chen, Chen, Chen, Chen, Chen, Chen, Chen, Chu, Dong, Duan, Fan, Fei, Gao, Ge, Gu, Gu, Gui, Guo, Guo, He, Hu, Huang, Jiang, Jiao, Jin, Lei, Li, Li, Li, Li, Li, Li, Liu, Liu, Hong, Liu, Liu, Liu, Lv, Lv, Lv, Ma, Ma, Ma, Ning, Ouyang, Qiu, Qu, Shang, Shao, Song, Song, Sui, Sun, Sun, Tang, Wang, Wang, Wang, Wang, Wang, Wang, Wang, Wei, Weng, Wu, Xiong, Xu, Xu, Yan, Yan, Yang, Ye, Ying, Yu, Yu, Zang, Zhang, Zhang, Zhang, Zhang, Zhang, Zhang, Zhang, Zhang, Zhang, Zhang, Zhang, Zhao, Zhao, Zhao, Zhou, Zhou, Zhuo, Zou, Qiu, Qiao, and Lin}]{internlm2}
Zheng Cai, Maosong Cao, Haojiong Chen, Kai Chen, Keyu Chen, Xin Chen, Xun Chen, Zehui Chen, Zhi Chen, Pei Chu, Xiaoyi Dong, Haodong Duan, Qi~Fan, Zhaoye Fei, Yang Gao, Jiaye Ge, Chenya Gu, Yuzhe Gu, Tao Gui, Aijia Guo, Qipeng Guo, Conghui He, Yingfan Hu, Ting Huang, Tao Jiang, Penglong Jiao, Zhenjiang Jin, Zhikai Lei, Jiaxing Li, Jingwen Li, Linyang Li, Shuaibin Li, Wei Li, Yining Li, Hongwei Liu, Jiangning Liu, Jiawei Hong, Kaiwen Liu, Kuikun Liu, Xiaoran Liu, Chengqi Lv, Haijun Lv, Kai Lv, Li~Ma, Runyuan Ma, Zerun Ma, Wenchang Ning, Linke Ouyang, Jiantao Qiu, Yuan Qu, Fukai Shang, Yunfan Shao, Demin Song, Zifan Song, Zhihao Sui, Peng Sun, Yu~Sun, Huanze Tang, Bin Wang, Guoteng Wang, Jiaqi Wang, Jiayu Wang, Rui Wang, Yudong Wang, Ziyi Wang, Xingjian Wei, Qizhen Weng, Fan Wu, Yingtong Xiong, Chao Xu, Ruiliang Xu, Hang Yan, Yirong Yan, Xiaogui Yang, Haochen Ye, Huaiyuan Ying, Jia Yu, Jing Yu, Yuhang Zang, Chuyu Zhang, Li~Zhang, Pan Zhang, Peng Zhang, Ruijie Zhang, Shuo Zhang, Songyang Zhang, Wenjian Zhang,
  Wenwei Zhang, Xingcheng Zhang, Xinyue Zhang, Hui Zhao, Qian Zhao, Xiaomeng Zhao, Fengzhe Zhou, Zaida Zhou, Jingming Zhuo, Yicheng Zou, Xipeng Qiu, Yu~Qiao, and Dahua Lin. 2024.
\newblock \href {https://arxiv.org/abs/2403.17297} {Internlm2 technical report}.
\newblock \emph{Preprint}, arXiv:2403.17297.

\bibitem[{Chen et~al.(2022)Chen, Deng, Wu, Gu, and Li}]{moe}
Zixiang Chen, Yihe Deng, Yue Wu, Quanquan Gu, and Yuanzhi Li. 2022.
\newblock \href {https://arxiv.org/abs/2208.02813} {Towards understanding mixture of experts in deep learning}.
\newblock \emph{Preprint}, arXiv:2208.02813.

\bibitem[{Cohan et~al.(2020)Cohan, Feldman, Beltagy, Downey, and Weld}]{specter}
Arman Cohan, Sergey Feldman, Iz~Beltagy, Doug Downey, and Daniel~S. Weld. 2020.
\newblock \href {https://arxiv.org/abs/2004.07180} {Specter: Document-level representation learning using citation-informed transformers}.
\newblock \emph{Preprint}, arXiv:2004.07180.

\bibitem[{Company(2012)}]{neo4j}
Neo4j Company. 2012.
\newblock \href {http://neo4j.org/} {Neo4j - the world's leading graph database}.

\bibitem[{Cormack et~al.(2009)Cormack, Clarke, and Buettcher}]{rrf}
Gordon~V Cormack, Charles~LA Clarke, and Stefan Buettcher. 2009.
\newblock Reciprocal rank fusion outperforms condorcet and individual rank learning methods.
\newblock In \emph{Proceedings of the 32nd international ACM SIGIR conference on Research and development in information retrieval}, pages 758--759.

\bibitem[{Dong et~al.(2023)Dong, Li, Dai, Zheng, Wu, Chang, Sun, Xu, Li, and Sui}]{icl}
Qingxiu Dong, Lei Li, Damai Dai, Ce~Zheng, Zhiyong Wu, Baobao Chang, Xu~Sun, Jingjing Xu, Lei Li, and Zhifang Sui. 2023.
\newblock \href {https://arxiv.org/abs/2301.00234} {A survey on in-context learning}.
\newblock \emph{Preprint}, arXiv:2301.00234.

\bibitem[{Dong et~al.(2022)Dong, Ni, Bikel, Alfonseca, Wang, Qu, and Zitouni}]{dea}
Zhe Dong, Jianmo Ni, Daniel~M. Bikel, Enrique Alfonseca, Yuan Wang, Chen Qu, and Imed Zitouni. 2022.
\newblock \href {https://arxiv.org/abs/2204.07120} {Exploring dual encoder architectures for question answering}.
\newblock \emph{Preprint}, arXiv:2204.07120.

\bibitem[{Du et~al.(2022)Du, Qian, Liu, Ding, Qiu, Yang, and Tang}]{glm}
Zhengxiao Du, Yujie Qian, Xiao Liu, Ming Ding, Jiezhong Qiu, Zhilin Yang, and Jie Tang. 2022.
\newblock \href {https://arxiv.org/abs/2103.10360} {Glm: General language model pretraining with autoregressive blank infilling}.
\newblock \emph{Preprint}, arXiv:2103.10360.

\bibitem[{Galileo Mark~Namata and Huang(2012)}]{pubmed}
Lise~Getoor Galileo Mark~Namata, Ben~London and Bert Huang. 2012.
\newblock Query-driven active surveying for collective classification.
\newblock In \emph{International Workshop on Mining and Learning with Graphs}, Edinburgh, Scotland.

\bibitem[{Gao et~al.(2024)Gao, Xiong, Gao, Jia, Pan, Bi, Dai, Sun, Wang, and Wang}]{ragsurvey}
Yunfan Gao, Yun Xiong, Xinyu Gao, Kangxiang Jia, Jinliu Pan, Yuxi Bi, Yi~Dai, Jiawei Sun, Meng Wang, and Haofen Wang. 2024.
\newblock \href {https://arxiv.org/abs/2312.10997} {Retrieval-augmented generation for large language models: A survey}.
\newblock \emph{Preprint}, arXiv:2312.10997.

\bibitem[{Guo et~al.(2020)Guo, Jin, Qiu, Zhang, Wipf, and Zhang}]{cyclegt}
Qipeng Guo, Zhijing Jin, Xipeng Qiu, Weinan Zhang, David Wipf, and Zheng Zhang. 2020.
\newblock \href {https://arxiv.org/abs/2006.04702} {Cyclegt: Unsupervised graph-to-text and text-to-graph generation via cycle training}.
\newblock \emph{Preprint}, arXiv:2006.04702.

\bibitem[{Hendrycks et~al.(2021)Hendrycks, Burns, Basart, Zou, Mazeika, Song, and Steinhardt}]{mmlu}
Dan Hendrycks, Collin Burns, Steven Basart, Andy Zou, Mantas Mazeika, Dawn Song, and Jacob Steinhardt. 2021.
\newblock \href {https://arxiv.org/abs/2009.03300} {Measuring massive multitask language understanding}.
\newblock \emph{Preprint}, arXiv:2009.03300.

\bibitem[{Himmelstein et~al.(2017)Himmelstein, Lizee, Hessler, Brueggeman, Chen, Hadley, Green, Khankhanian, and Baranzini}]{hetionet}
Daniel~Scott Himmelstein, Antoine Lizee, Christine Hessler, Leo Brueggeman, Sabrina~L Chen, Dexter Hadley, Ari Green, Pouya Khankhanian, and Sergio~E Baranzini. 2017.
\newblock \href {https://doi.org/10.7554/eLife.26726} {Systematic integration of biomedical knowledge prioritizes drugs for repurposing}.
\newblock \emph{eLife}, 6:e26726.

\bibitem[{Huang et~al.(2024)Huang, Ping, Xu, Shoeybi, Chang, and Catanzaro}]{raven}
Jie Huang, Wei Ping, Peng Xu, Mohammad Shoeybi, Kevin Chen-Chuan Chang, and Bryan Catanzaro. 2024.
\newblock \href {https://arxiv.org/abs/2308.07922} {Raven: In-context learning with retrieval-augmented encoder-decoder language models}.
\newblock \emph{Preprint}, arXiv:2308.07922.

\bibitem[{Izacard et~al.(2022{\natexlab{a}})Izacard, Caron, Hosseini, Riedel, Bojanowski, Joulin, and Grave}]{contriever}
Gautier Izacard, Mathilde Caron, Lucas Hosseini, Sebastian Riedel, Piotr Bojanowski, Armand Joulin, and Edouard Grave. 2022{\natexlab{a}}.
\newblock \href {https://arxiv.org/abs/2112.09118} {Unsupervised dense information retrieval with contrastive learning}.
\newblock \emph{Preprint}, arXiv:2112.09118.

\bibitem[{Izacard et~al.(2022{\natexlab{b}})Izacard, Lewis, Lomeli, Hosseini, Petroni, Schick, Dwivedi-Yu, Joulin, Riedel, and Grave}]{atlas}
Gautier Izacard, Patrick Lewis, Maria Lomeli, Lucas Hosseini, Fabio Petroni, Timo Schick, Jane Dwivedi-Yu, Armand Joulin, Sebastian Riedel, and Edouard Grave. 2022{\natexlab{b}}.
\newblock \href {https://arxiv.org/abs/2208.03299} {Atlas: Few-shot learning with retrieval augmented language models}.
\newblock \emph{Preprint}, arXiv:2208.03299.

\bibitem[{Jiang et~al.(2023)Jiang, Xu, Gao, Sun, Liu, Dwivedi-Yu, Yang, Callan, and Neubig}]{flare}
Zhengbao Jiang, Frank~F. Xu, Luyu Gao, Zhiqing Sun, Qian Liu, Jane Dwivedi-Yu, Yiming Yang, Jamie Callan, and Graham Neubig. 2023.
\newblock \href {https://arxiv.org/abs/2305.06983} {Active retrieval augmented generation}.
\newblock \emph{Preprint}, arXiv:2305.06983.

\bibitem[{Jin et~al.(2020)Jin, Pan, Oufattole, Weng, Fang, and Szolovits}]{textbooks}
Di~Jin, Eileen Pan, Nassim Oufattole, Wei-Hung Weng, Hanyi Fang, and Peter Szolovits. 2020.
\newblock \href {https://arxiv.org/abs/2009.13081} {What disease does this patient have? a large-scale open domain question answering dataset from medical exams}.
\newblock \emph{Preprint}, arXiv:2009.13081.

\bibitem[{Jin et~al.(2019)Jin, Dhingra, Liu, Cohen, and Lu}]{pubmedqa}
Qiao Jin, Bhuwan Dhingra, Zhengping Liu, William~W. Cohen, and Xinghua Lu. 2019.
\newblock \href {https://arxiv.org/abs/1909.06146} {Pubmedqa: A dataset for biomedical research question answering}.
\newblock \emph{Preprint}, arXiv:1909.06146.

\bibitem[{Jin et~al.(2023)Jin, Kim, Chen, Comeau, Yeganova, Wilbur, and Lu}]{medcpt}
Qiao Jin, Won Kim, Qingyu Chen, Donald~C Comeau, Lana Yeganova, W~John Wilbur, and Zhiyong Lu. 2023.
\newblock \href {https://doi.org/10.1093/bioinformatics/btad651} {Medcpt: Contrastive pre-trained transformers with large-scale pubmed search logs for zero-shot biomedical information retrieval}.
\newblock \emph{Bioinformatics}, 39(11).

\bibitem[{Johnson et~al.(2017)Johnson, Douze, and Jégou}]{faiss}
Jeff Johnson, Matthijs Douze, and Hervé Jégou. 2017.
\newblock \href {https://arxiv.org/abs/1702.08734} {Billion-scale similarity search with gpus}.
\newblock \emph{Preprint}, arXiv:1702.08734.

\bibitem[{Kang et~al.(2023)Kang, Kwak, Baek, and Hwang}]{surge}
Minki Kang, Jin~Myung Kwak, Jinheon Baek, and Sung~Ju Hwang. 2023.
\newblock \href {https://arxiv.org/abs/2305.18846} {Knowledge graph-augmented language models for knowledge-grounded dialogue generation}.
\newblock \emph{Preprint}, arXiv:2305.18846.

\bibitem[{Ke et~al.(2024)Ke, Jin, Elangovan, Abdullah, Liu, Sia, Soh, Tung, Ong, and Ting}]{rag4healthcare}
YuHe Ke, Liyuan Jin, Kabilan Elangovan, Hairil~Rizal Abdullah, Nan Liu, Alex Tiong~Heng Sia, Chai~Rick Soh, Joshua Yi~Min Tung, Jasmine Chiat~Ling Ong, and Daniel Shu~Wei Ting. 2024.
\newblock \href {https://arxiv.org/abs/2402.01733} {Development and testing of retrieval augmented generation in large language models -- a case study report}.
\newblock \emph{Preprint}, arXiv:2402.01733.

\bibitem[{Khan(2023)}]{ragproducts}
Assia Khan. 2023.
\newblock Retrieval augmented generation: 5 uses and their examples.
\newblock \emph{Lettria}.

\bibitem[{Khandelwal et~al.(2020)Khandelwal, Levy, Jurafsky, Zettlemoyer, and Lewis}]{knnlm}
Urvashi Khandelwal, Omer Levy, Dan Jurafsky, Luke Zettlemoyer, and Mike Lewis. 2020.
\newblock \href {https://arxiv.org/abs/1911.00172} {Generalization through memorization: Nearest neighbor language models}.
\newblock \emph{Preprint}, arXiv:1911.00172.

\bibitem[{Khattab et~al.(2023)Khattab, Santhanam, Li, Hall, Liang, Potts, and Zaharia}]{dsp}
Omar Khattab, Keshav Santhanam, Xiang~Lisa Li, David Hall, Percy Liang, Christopher Potts, and Matei Zaharia. 2023.
\newblock \href {https://arxiv.org/abs/2212.14024} {Demonstrate-search-predict: Composing retrieval and language models for knowledge-intensive nlp}.
\newblock \emph{Preprint}, arXiv:2212.14024.

\bibitem[{Lewis et~al.(2020)Lewis, Perez, Piktus, Petroni, Karpukhin, Goyal, K\"{u}ttler, Lewis, Yih, Rockt\"{a}schel, Riedel, and Kiela}]{rag}
Patrick Lewis, Ethan Perez, Aleksandra Piktus, Fabio Petroni, Vladimir Karpukhin, Naman Goyal, Heinrich K\"{u}ttler, Mike Lewis, Wen-tau Yih, Tim Rockt\"{a}schel, Sebastian Riedel, and Douwe Kiela. 2020.
\newblock \href {https://proceedings.neurips.cc/paper_files/paper/2020/file/6b493230205f780e1bc26945df7481e5-Paper.pdf} {Retrieval-augmented generation for knowledge-intensive nlp tasks}.
\newblock In \emph{Advances in Neural Information Processing Systems}, volume~33, pages 9459--9474. Curran Associates, Inc.

\bibitem[{Liang et~al.(2024)Liang, Zhang, Li, Yu, and Xu}]{intergen}
Han Liang, Wenqian Zhang, Wenxuan Li, Jingyi Yu, and Lan Xu. 2024.
\newblock \href {https://doi.org/10.1007/s11263-024-02042-6} {Intergen: Diffusion-based multi-human motion generation under complex interactions}.
\newblock \emph{International Journal of Computer Vision}.

\bibitem[{Liu et~al.(2019)Liu, Ott, Goyal, Du, Joshi, Chen, Levy, Lewis, Zettlemoyer, and Stoyanov}]{roberta}
Yinhan Liu, Myle Ott, Naman Goyal, Jingfei Du, Mandar Joshi, Danqi Chen, Omer Levy, Mike Lewis, Luke Zettlemoyer, and Veselin Stoyanov. 2019.
\newblock \href {https://arxiv.org/abs/1907.11692} {Roberta: A robustly optimized bert pretraining approach}.
\newblock \emph{Preprint}, arXiv:1907.11692.

\bibitem[{Melnyk et~al.(2022)Melnyk, Dognin, and Das}]{text2kg}
Igor Melnyk, Pierre Dognin, and Payel Das. 2022.
\newblock \href {https://arxiv.org/abs/2211.10511} {Knowledge graph generation from text}.
\newblock \emph{Preprint}, arXiv:2211.10511.

\bibitem[{Nishikawa et~al.(2022)Nishikawa, Ri, Yamada, Tsuruoka, and Echizen}]{ease}
Sosuke Nishikawa, Ryokan Ri, Ikuya Yamada, Yoshimasa Tsuruoka, and Isao Echizen. 2022.
\newblock \href {https://arxiv.org/abs/2205.04260} {Ease: Entity-aware contrastive learning of sentence embedding}.
\newblock \emph{Preprint}, arXiv:2205.04260.

\bibitem[{Pal et~al.(2022)Pal, Umapathi, and Sankarasubbu}]{medmcqa}
Ankit Pal, Logesh~Kumar Umapathi, and Malaikannan Sankarasubbu. 2022.
\newblock \href {https://arxiv.org/abs/2203.14371} {Medmcqa : A large-scale multi-subject multi-choice dataset for medical domain question answering}.
\newblock \emph{Preprint}, arXiv:2203.14371.

\bibitem[{Publishing(2024)}]{statpearls}
StatPearls Publishing. 2024.
\newblock Statpearls.
\newblock Treasure Island (FL): StatPearls Publishing.
\newblock Available from: \url{https://www.ncbi.nlm.nih.gov/books/NBK430685/}.

\bibitem[{Radeva et~al.(2024)Radeva, Popchev, Doukovska, and Dimitrova}]{passer}
Irina Radeva, Ivan Popchev, Lyubka Doukovska, and Miroslava Dimitrova. 2024.
\newblock \href {https://doi.org/10.3390/electronics13071361} {Web application for retrieval-augmented generation: Implementation and testing}.
\newblock \emph{Electronics}, 13:1361.

\bibitem[{Ram et~al.(2023)Ram, Levine, Dalmedigos, Muhlgay, Shashua, Leyton-Brown, and Shoham}]{ralm}
Ori Ram, Yoav Levine, Itay Dalmedigos, Dor Muhlgay, Amnon Shashua, Kevin Leyton-Brown, and Yoav Shoham. 2023.
\newblock \href {https://arxiv.org/abs/2302.00083} {In-context retrieval-augmented language models}.
\newblock \emph{Preprint}, arXiv:2302.00083.

\bibitem[{Ramos(2003)}]{tfidf}
Juan~Enrique Ramos. 2003.
\newblock \href {https://api.semanticscholar.org/CorpusID:14638345} {Using {TF-IDF} to determine word relevance in document queries}.

\bibitem[{Robertson and Zaragoza(2009)}]{bm25}
Stephen Robertson and Hugo Zaragoza. 2009.
\newblock \href {https://doi.org/10.1561/1500000019} {The probabilistic relevance framework: Bm25 and beyond}.
\newblock \emph{Found. Trends Inf. Retr.}, 3(4):333–389.

\bibitem[{Safjan(2023)}]{slidingwindow}
Krystian Safjan. 2023.
\newblock From fixed-size to nlp chunking - a deep dive into text chunking techniques.
\newblock \emph{Krystian's Safjan Blog}.

\bibitem[{Sarthi et~al.(2024)Sarthi, Abdullah, Tuli, Khanna, Goldie, and Manning}]{raptor}
Parth Sarthi, Salman Abdullah, Aditi Tuli, Shubh Khanna, Anna Goldie, and Christopher~D. Manning. 2024.
\newblock \href {https://arxiv.org/abs/2401.18059} {Raptor: Recursive abstractive processing for tree-organized retrieval}.
\newblock \emph{Preprint}, arXiv:2401.18059.

\bibitem[{Shi et~al.(2023)Shi, Min, Yasunaga, Seo, James, Lewis, Zettlemoyer, and tau Yih}]{replug}
Weijia Shi, Sewon Min, Michihiro Yasunaga, Minjoon Seo, Rich James, Mike Lewis, Luke Zettlemoyer, and Wen tau Yih. 2023.
\newblock \href {https://arxiv.org/abs/2301.12652} {Replug: Retrieval-augmented black-box language models}.
\newblock \emph{Preprint}, arXiv:2301.12652.

\bibitem[{Siegler(2024)}]{metadatafiltering}
Ryan Siegler. 2024.
\newblock Optimizing vector search with metadata filtering.
\newblock \emph{KX systems}.

\bibitem[{Soman and Roychowdhury(2024)}]{ragtechdoc}
Sumit Soman and Sujoy Roychowdhury. 2024.
\newblock \href {https://arxiv.org/abs/2404.00657} {Observations on building rag systems for technical documents}.
\newblock \emph{Preprint}, arXiv:2404.00657.

\bibitem[{team(2024)}]{cec}
Antematter team. 2024.
\newblock Optimizing retrieval-augmented generation with advanced chunking techniques: A comparative study.
\newblock \emph{Antematter}.

\bibitem[{team(2023)}]{pdr}
LangChain team. 2023.
\newblock Parent document retriever.
\newblock \emph{LangChain document}.

\bibitem[{Tsatsaronis et~al.(2015)Tsatsaronis, Balikas, Malakasiotis, Partalas, Zschunke, Alvers, Weißenborn, Krithara, Petridis, Polychronopoulos, Almirantis, Pavlopoulos, Baskiotis, Gallinari, Artieres, Ngonga~Ngomo, Heino, Gaussier, Barrio-Alvers, and Paliouras}]{bioasq}
George Tsatsaronis, Georgios Balikas, Prodromos Malakasiotis, Ioannis Partalas, Matthias Zschunke, Michael Alvers, Dirk Weißenborn, Anastasia Krithara, Sergios Petridis, Dimitris Polychronopoulos, Yannis Almirantis, John Pavlopoulos, Nicolas Baskiotis, Patrick Gallinari, Thierry Artieres, Axel-Cyrille Ngonga~Ngomo, Norman Heino, Eric Gaussier, Liliana Barrio-Alvers, and Georgios Paliouras. 2015.
\newblock \href {https://doi.org/10.1186/s12859-015-0564-6} {An overview of the bioasq large-scale biomedical semantic indexing and question answering competition}.
\newblock \emph{BMC Bioinformatics}, 16:138.

\bibitem[{Wang et~al.(2023{\natexlab{a}})Wang, Jiang, You, Han, Liu, Srinivasa, Kompella, and Wang}]{gmoe}
Haotao Wang, Ziyu Jiang, Yuning You, Yan Han, Gaowen Liu, Jayanth Srinivasa, Ramana~Rao Kompella, and Zhangyang Wang. 2023{\natexlab{a}}.
\newblock \href {https://arxiv.org/abs/2304.02806} {Graph mixture of experts: Learning on large-scale graphs with explicit diversity modeling}.
\newblock \emph{Preprint}, arXiv:2304.02806.

\bibitem[{Wang et~al.(2023{\natexlab{b}})Wang, Li, Sun, and Liu}]{skr}
Yile Wang, Peng Li, Maosong Sun, and Yang Liu. 2023{\natexlab{b}}.
\newblock \href {https://arxiv.org/abs/2310.05002} {Self-knowledge guided retrieval augmentation for large language models}.
\newblock \emph{Preprint}, arXiv:2310.05002.

\bibitem[{Wei et~al.(2022)Wei, Wang, Schuurmans, Bosma, ichter, Xia, Chi, Le, and Zhou}]{cot}
Jason Wei, Xuezhi Wang, Dale Schuurmans, Maarten Bosma, brian ichter, Fei Xia, Ed~Chi, Quoc~V Le, and Denny Zhou. 2022.
\newblock \href {https://proceedings.neurips.cc/paper_files/paper/2022/file/9d5609613524ecf4f15af0f7b31abca4-Paper-Conference.pdf} {Chain-of-thought prompting elicits reasoning in large language models}.
\newblock In \emph{Advances in Neural Information Processing Systems}, volume~35, pages 24824--24837. Curran Associates, Inc.

\bibitem[{{Wikimedia Foundation}(2024)}]{wikipedia_data}
{Wikimedia Foundation}. 2024.
\newblock Wikipedia.
\newblock The Free Encyclopedia.
\newblock Available from: \url{https://www.wikipedia.org}.

\bibitem[{Xiong et~al.(2024)Xiong, Jin, Lu, and Zhang}]{medrag}
Guangzhi Xiong, Qiao Jin, Zhiyong Lu, and Aidong Zhang. 2024.
\newblock \href {https://arxiv.org/abs/2402.13178} {Benchmarking retrieval-augmented generation for medicine}.
\newblock \emph{Preprint}, arXiv:2402.13178.

\bibitem[{You et~al.(2018)You, Ying, Ren, Hamilton, and Leskovec}]{graphrnn}
Jiaxuan You, Rex Ying, Xiang Ren, William~L. Hamilton, and Jure Leskovec. 2018.
\newblock \href {https://arxiv.org/abs/1802.08773} {Graphrnn: Generating realistic graphs with deep auto-regressive models}.
\newblock \emph{Preprint}, arXiv:1802.08773.

\bibitem[{Zhang et~al.(2018)Zhang, Wu, He, Liu, and Su}]{medqa}
Xiao Zhang, Ji~Wu, Zhiyang He, Xien Liu, and Ying Su. 2018.
\newblock \href {https://arxiv.org/abs/1802.10279} {Medical exam question answering with large-scale reading comprehension}.
\newblock \emph{Preprint}, arXiv:1802.10279.

\end{thebibliography}

\clearpage
\appendix

\section{More Discussion About Selection Process}
\label{sec:discussion_selection_process}
The intuitive selection process would be to select the top snippets with the highest weighted similarity scores. However, this method is not adopted for the following reasons:

(1) It is biased, because more-coarsed snippets tend to have higher similarity scores, even though they might include more noise. 

(2) It is not rebust to rely on the router's output weight to control such imbalance between granularity levels, because the router is not traine for this objective. The router is trained with soft labels, which are approximate training signals that can only teach the model to distinguish the 1st and 2nd optimal granularity level from the rest.

(3) In practice, when $\mathcal{K}$ is too large, it has to be stored in separated vector databases. A common solution is use one vectorbase to store one granularity. This solution will make the similarity scores across granularity levels NOT directly comparable with each other.

Based on these reasons, we designed the selection strategy presented in the paper. The rationale behind the proposed strategy is that: Since we aim to extract the globally optimal snippet, we need a way to calculate a ``global highest similarity score''. Given that the finest-grained snippets are the only ones that appear across all granularity levels following our initial chunking approach (fixed-size, non-overlapping), we calculate the weighted sum based on these finest-grained snippets.

\section{Details of QA Datasets}
\label{sec:qa_datasets}
\subsection{MMLU-Med}
The Massive Multitask Language Understanding (MMLU) benchmark \cite{mmlu} evaluates the multitask learning capability of language models. While the full MMLU dataset encompasses 57 different tasks, we specifically extracted the medical questions for our tests, totaling 1089 questions.

\subsection{MedQA-US}
MedQA \cite{medqa} is a multiple-choice QA dataset derived from professional medical board exams. It is available in Simplified Chinese, Traditional Chinese, and English. For our experiments, we used 1273 questions from the English version.

\subsection{MedMCQA}
MedMCQA \cite{medmcqa} comprises a large number of questions from the Indian medical entrance exam, covering 2400 healthcare topics and 21 medical subjects. For the MIRAGE benchmark, we utilized the ``dev'' set of the original MedMCQA dataset.

\subsection{PubMedQA*}
PubMedQA is a research QA dataset in the biomedical field, consisting of 1000 manually annotated questions constructed from PubMed abstracts. In the MIRAGE benchmark, the reference contexts were removed. We selected a subset of 500 questions, which we refer to as PubMedQA*.

\subsection{BioASQ-Y/N}
BioASQ \cite{bioasq} is an annual biomedical QA competition. For the MIRAGE benchmark, we selected only the Machine Reading Comprehensive Track (Task B), focusing on 618 questions from recent years (2019-2023).

The important statistics of the copora and the QA datasets are presente in Table \ref{tab:corpus_dataset_table} below.

\begin{table*}[ht]
  \centering
  \footnotesize
  \begin{tabular}{C{1.2cm}C{0.8cm}C{0.8cm}C{0.4cm}C{1.4cm}}
    \toprule
    Corpus  & \#Doc & \#Snip. & $\bar{L}$. & Domain\\
    \midrule
    PubMed & 23.9M  & 23.9M & 196 & Bio.-Med.     \\
    StatPearls & 9.3k  & 301.2k & 119 & Clinics     \\
    Textbooks & 18 & 125.8k & 182 & Medicine     \\
    Wikipedia & 6.5M  & 29.9M & 162 & General     \\
    MedCorp & 30.4M  & 54.2M & 221 & Mixed     \\
    \bottomrule
  \end{tabular}
\quad
  \footnotesize
  \begin{tabular}{C{1.8cm}C{0.7cm}C{0.4cm}C{0.5cm}C{1.4cm}}
    \toprule
    Datasets  & Size &\#Opt. & $\bar{L}$ & Source\\
    \midrule
    MMLU-Med & 1,089  & 4 & 63 & Exam.     \\
    MedQA-US & 1,273  & 4 & 177 & Exam.     \\
    MedMCQA & 4,183  & 4 & 26 & Exam.     \\
    PubMedQA* & 500  & 3 & 24 & Literature     \\
    BioASQ-Y/N & 618  & 2 & 17 & Literature     \\
    \bottomrule
  \end{tabular}
\caption{Important statistics of corpora and QA datasets}
\label{tab:corpus_dataset_table}
\end{table*}

\section{Exact Versions of LLMs}
\label{sec:llm_versions}
The exact versions of the backbone LLMs tested are listed in Table \ref{tab:llm_version_table}.
\begin{table*}[hb]
\centering
    \small
  \begin{tabular}{ccl}
    \toprule
    LLM name  & LLM version & LLM site \\
    \midrule
    ChatGPT & gpt-3.5-turbo-16k  & https://platform.openai.com/docs/models/gpt-3.5-turbo   \\
    Llama3 & Meta-Llama-3-8B  & https://huggingface.co/meta-llama/Meta-Llama-3-8B-Instruct  \\
    InternLM2 & internlm2-123b & https://www.sensetime.com/en/news-detail/51167237 \\
    ChatGLM3 & chatglm3-6b  & https://huggingface.co/THUDM/chatglm3-6b  \\
    Qwen1.5 & Qwen1.5-MoE-A2.7B   & https://huggingface.co/Qwen/Qwen1.5-MoE-A2.7B-Chat     \\
    \bottomrule
  \end{tabular}
  \caption{Versions of the backbone LLMs}
  \label{tab:llm_version_table}
\end{table*}

\section{Qualitative Results of the Router}
\label{sec:qualitative_res}
As illustrated in Figure \ref{fig:granularity distribution}, the router effectively assigns different weights to various granularity levels. From the figure, we can infer that a potential global peak in weights may exist at a granularity level smaller than level 1 or larger than level 5.

However, we did not test these smaller or larger granularity levels due to the following reasons: In our experiments, we focus exclusively on LLMs with a parameter size of around 7 billion. For these models, the chunking size at granularity level 5 approaches their maximum context window. Conversely, the chunking size at granularity level 1 consists of only a few dozen characters, which is already quite small.

\section{Experiment on Soft Labels}
\label{sec:soft_labels}
In this section, we evaluate the performance of building soft labels using different methods (TF-IDF \cite{tfidf}, RoBERTa \cite{roberta}, and hitrate score). For this experiment, we fix one retriever (BM25) and the backbone LLM (Qwen1.5), then build the soft labels using these three different methods. Different routers are trained with these different soft labels and tested on various MQA datasets. The results are grouped in Table \ref{tab:soft_label_table}. 

The results indicate that there is no universal best method for building soft labels. For instance, ``hitrate score'' is the most effective when training on MedMCQA, while RoBERTa shows advantages on PubMedQA. Overall, the soft labels built with RoBERTa demonstrate good performance across the board. Therefore, for the remainder of the experiments, RoBERTa is adopted as the default method to build soft labels.

\begin{table*}[ht]

  \centering
  \small
  \begin{tabular}{C{1.4cm}C{1.3cm}C{1.4cm}C{1.4cm}C{1.4cm}C{1.4cm}C{1.4cm}C{1cm}}
  \toprule
\multirow{2}{*}{\shortstack{Training\\datasets}}& \multirow{2}{*}{Methods}  & \multicolumn{6}{c}{MIRAGE Benchmark Dataset (Acc.)}     \\
                        &                                                   & \small{MMLU}  & \small{MedQA} & \small{MedMCQA}   & \small{PubMedQA}  & \small{BioASQ}    & \small{Avg.}    \\
\midrule
\multirow{3}{*}{\small{MedQA}}     & RoBERTa                                     & 0.4265\textcolor{g}{$\pm$0.04}          & 0.3711\textcolor{g}{$\pm$0.04}            & 0.3831\textcolor{g}{$\pm$0.02}                                                                                                                                      & \textbf{0.5000}\textcolor{g}{$\pm$0.06} & 0.6234\textcolor{g}{$\pm$0.06}            & 0.4608 \\
                                   & \textbf{hitrate}                            & \textbf{0.4779}\textcolor{g}{$\pm$0.04} & \textbf{0.4088}\textcolor{g}{$\pm$0.04}   & \textbf{0.4330}\textcolor{g}{$\pm$0.02}                                                                                          & 0.4194\textcolor{g}{$\pm$0.06}          & \textbf{0.6753}\textcolor{g}{$\pm$0.05}   & \textbf{0.4829} \\
                                   & TF-IDF                                      & 0.4559\textcolor{g}{$\pm$0.04}          & 0.3019\textcolor{g}{$\pm$0.04}            & 0.4023\textcolor{g}{$\pm$0.02}                                                                                                   & 0.4355\textcolor{g}{$\pm$0.06}          & 0.5714\textcolor{g}{$\pm$0.06}            & 0.4334 \\
\midrule
\multirow{3}{*}{\small{MedMCQA}}   & RoBERTa                                     & \textbf{0.4485}\textcolor{g}{$\pm$0.05} & 0.3459\textcolor{g}{$\pm$0.04}            & 0.4023\textcolor{g}{$\pm$0.02}                                                                                                                                      & 0.4839\textcolor{g}{$\pm$0.06}          & 0.5195\textcolor{g}{$\pm$0.06}            & 0.4400 \\
                                   & hitrate                                     & 0.4118\textcolor{g}{$\pm$0.04}          & 0.3774\textcolor{g}{$\pm$0.04}            & 0.3678\textcolor{g}{$\pm$0.02}                                                                                                   & 0.3387\textcolor{g}{$\pm$0.06}          & 0.5065\textcolor{g}{$\pm$0.06}            & 0.4004 \\
                                   & \textbf{TF-IDF}                             & 0.4412\textcolor{g}{$\pm$0.04}          & \textbf{0.3899}\textcolor{g}{$\pm$0.04}   & \textbf{0.4119}\textcolor{g}{$\pm$0.02}                                                                                          & \textbf{0.5161}\textcolor{g}{$\pm$0.06} &\textbf{0.5844}\textcolor{g}{$\pm$0.06}    & \textbf{0.4687} \\
\midrule
\multirow{3}{*}{\small{PubMedQA}}  & \textbf{RoBERTa}                            & \textbf{0.4485}\textcolor{g}{$\pm$0.04} & \textbf{0.4528}\textcolor{g}{$\pm$0.04}   & \textbf{0.3966}\textcolor{g}{$\pm$0.02}                                                                                                                             & \textbf{0.4677}\textcolor{g}{$\pm$0.06} & \textbf{0.6623}\textcolor{g}{$\pm$0.05}   & \textbf{0.4856} \\
                                   & hitrate                                     & 0.4118\textcolor{g}{$\pm$0.04}          & 0.4088\textcolor{g}{$\pm$0.04}            & 0.3851\textcolor{g}{$\pm$0.02}                                                                                                   & 0.4355\textcolor{g}{$\pm$0.06}          & 0.6104\textcolor{g}{$\pm$0.06}            & 0.4503 \\
                                   & TF-IDF                                      & 0.4044\textcolor{g}{$\pm$0.04}          & 0.3208\textcolor{g}{$\pm$0.04}            & 0.3908\textcolor{g}{$\pm$0.02}                                                                                                   & \textbf{0.4677}\textcolor{g}{$\pm$0.06} & 0.5584\textcolor{g}{$\pm$0.06}            & 0.4284 \\
\bottomrule

\end{tabular} 
  
  \caption{The performance of the routers trained with the soft labels created with different methods, best method marked in \textbf{bold}.}
  \label{tab:soft_label_table}
\end{table*}

\section{Impact of the Number of Candidate Snippets}
\label{sec:rag_k}
In the previous experiment, three candidate snippets were retrieved from each granularity level of the external knowledge database. In this section, we investigate the effect of varying the number of candidate snippets on the overall performance of the RAG system. The rationale behind this exploration lies in the potential limitation of a small pool of candidate snippets, such as keeping only three. In such case, valuable snippets may be overlooked. For instance, a snippet ranked fourth or fifth in each retrieval may appear repeatedly across different granularity levels and thus might be selected after its relevance scores adjusted with weights assigned by the router. However, despite potentially high relevance, such snippets are excluded early in the retrieval process.

\begin{figure*}[ht]
  \centering
  \includegraphics[width=1\linewidth]{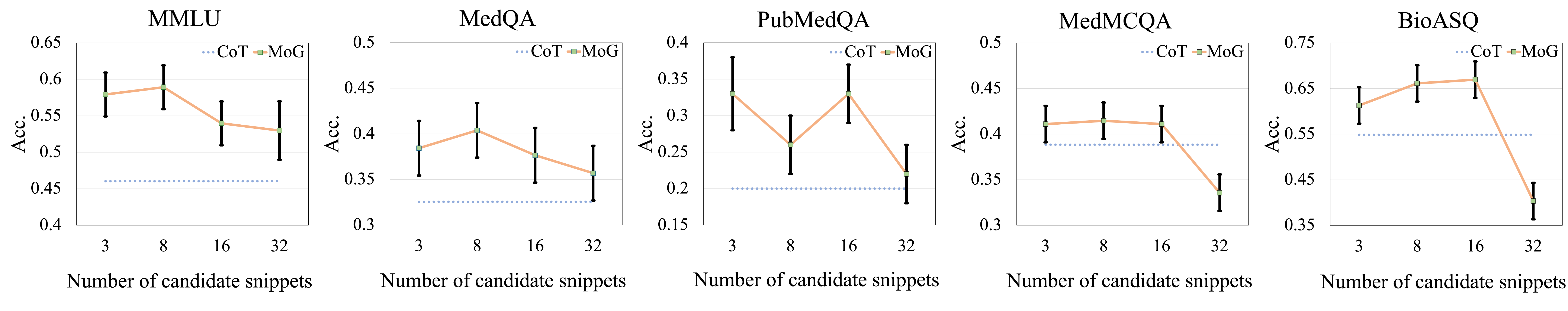}
  \caption{Accuracy of Medical Question-Answering task with different number of candidate snippets}
  \label{fig:number_candidate_snippet}
\end{figure*}

In this experiment, we test the RAG system equipped with MoG using different numbers of candidate snippets $k_r$ ($k_r \in $ \{3, 8, 16, 32\}). For simplicity, the backbone LLM was fixed as Qwen1.5. The experiment results are shown in Figure \ref{fig:number_candidate_snippet}. 

Generally, RAG performance improves as $k_r$ increases, indicating the presence of helpful knowledge in the retrieved snippets. However, too many irrelevant snippets mislead the inference of LLM based on its knowledge. MoG stands out with its high initial performance even at low $k_r$ values ($k_r$<=8 in this case), thanks to its multi-granularity filtering, which efficiently selects relevant snippets. This enhances the LLM's accuracy without requiring a large snippet pool. MoG even strikes a balance at high $k_r$ values ($k_r$>=16 in this case) due to its threshold limiting of the different corpora, avoiding the pitfalls of excessive information retrieval. MoG's consistently high performance demonstrates its superiority in optimizing the number of candidate snippets for effective medical question-answering.

\section{Choice of Retriever}
\label{sec:choice_retriever}
In the previous experiment, BM25 \cite{bm25} was used as the retriever because it is a lightweight and popular choice in practice. In this section, we replace BM25 with other popular retrievers to evaluate their performance. Referring to the setup in MedRAG \cite{medrag}, we select a general-domain semantic retriever called Contriever \cite{contriever}, a scientific-domain retriever called SPECTER \cite{specter}, and a biomedical-domain retriever called MedCPT \cite{medcpt}. Additionally, the Reciprocal Rank Fusion (RRF) method \cite{rrf} is utilized to combine results from different retrievers, including RRF-2 (fusion of BM25 and MedCPT) and RRF-4 (fusion of all four retrievers). From Table \ref{tab:retriever}, we can observe the following: Each retriever has a well-performing QA set, which results in minimal overall differences among these retrievers. Considering these conclusions, we decide to continue using BM25 for all other experiments.

\begin{table*}[h]
\small
\centering
\begin{tabular}{C{1.4cm}C{1.3cm}C{1.4cm}C{1.4cm}C{1.4cm}C{1.4cm}C{1.4cm}C{1cm}}
\toprule
\multirow{2}{*}{Corpus} & \multirow{2}{*}{Retriever} & \multicolumn{6}{c}{MIRAGE Benchmark Dataset (Acc.)} \\
                        &                            & MMLU  & MedQA  & MedMCQA & PubMedQA & BioASQ & Avg. \\

\midrule
\multirow{6}{*}{Textbooks}  & BM25   & 0.6139\textcolor{g}{$\pm$0.04} & 0.3961\textcolor{g}{$\pm$0.04}   & 0.3907\textcolor{g}{$\pm$0.02}   & 0.3100\textcolor{g}{$\pm$0.06} & 0.5000\textcolor{g}{$\pm$0.05}   & 0.4421 \\
                        & Contriever   & 0.5644\textcolor{g}{$\pm$0.03} & 0.4235\textcolor{g}{$\pm$0.03}   & 0.4074\textcolor{g}{$\pm$0.01}   & 0.2500\textcolor{g}{$\pm$0.04} & 0.5161\textcolor{g}{$\pm$0.04}   & 0.4323 \\
                        & SPECTER   & 0.5941\textcolor{g}{$\pm$0.03} & 0.4118\textcolor{g}{$\pm$0.03}   & 0.3919\textcolor{g}{$\pm$0.01}   & 0.1900\textcolor{g}{$\pm$0.04} & 0.4839\textcolor{g}{$\pm$0.04}   & 0.4143 \\
                        & MedCPT   & 0.5347\textcolor{g}{$\pm$0.04} & 0.4549\textcolor{g}{$\pm$0.03}   & 0.4026\textcolor{g}{$\pm$0.02}   & 0.3100\textcolor{g}{$\pm$0.05} & 0.5565\textcolor{g}{$\pm$0.04}   & 0.4517 \\
                        & RRF-2   & 0.5396\textcolor{g}{$\pm$0.03} & 0.4196\textcolor{g}{$\pm$0.03}   & 0.4241\textcolor{g}{$\pm$0.02}   & 0.2800\textcolor{g}{$\pm$0.04} & 0.5403\textcolor{g}{$\pm$0.04}   & 0.4407 \\
                        & RRF-4   & 0.5495\textcolor{g}{$\pm$0.04} & 0.4392\textcolor{g}{$\pm$0.03}   & 0.4397\textcolor{g}{$\pm$0.02}   & 0.2600\textcolor{g}{$\pm$0.04} & 0.6129\textcolor{g}{$\pm$0.04}   & 0.4603 \\
\bottomrule
\end{tabular}
\caption{Performance of different retrievers}
\label{tab:retriever}
\end{table*}

\section{Performance of MoG on Medical QA Task (Other results)}
\label{sec:mog_other_results}
In this section, we present the experiment result of MoG on the Medical QA task with four different corpora as the training datasets for the router. The results are grouped in Table \ref{tab:mog_result_table_single_corpus}.

\begin{table*}[ht]
  \centering
  \small
  \begin{tabular}{C{1cm}C{1.1cm}C{1.5cm}C{1.5cm}C{1.5cm}C{1.5cm}C{1.5cm}} 
  \toprule
\multirow{2}{*}{LLM}        & \multirow{2}{*}{Method}                                                        & \multicolumn{4}{c}{Training Corpora (Avg Acc.)} \\

\cmidrule(r){3-6}
                            &                        & \small{Textbooks}                  & \small{StatPearls}                  & \small{PubMed}                   & \small{Wikipedia}                                      \\
   \midrule
\multirow{3}{*}{GLM3}       & CoT                    &0.3908                              &0.3908                              &0.3908                              &0.3908                                                                      \\
                            & MedRAG                 &0.4269                              &0.4343                              &0.4716                              &\textbf{0.4582}                                                                       \\
                            & MoG                    &\textbf{0.4492}                              &\textbf{0.4465}                              &\textbf{0.4911}                              &0.4241                                                                       \\
                        
   \midrule
\multirow{3}{*}{GPT-3.5}    & CoT                    &\textbf{0.6571}                              &\textbf{0.6571}                              &0.6571                              &\textbf{0.6571}                                                                       \\
                            & MedRAG                 &0.5250                              &0.5229                              &0.6322                              &0.5332                                                                       \\
                            & MoG                    &0.6122                              &0.5921                              &\textbf{0.6795}                              &0.6154                                                                       \\
                          
   \midrule
\multirow{3}{*}{InternLM}    & CoT                   &\textbf{0.5914}                              &\textbf{0.5914}                              &0.5914                              &\textbf{0.5914}                                                                       \\
                             & MedRAG                &0.4223                              &0.4380                              &0.5559                              &0.4506                                                                       \\
                             & MoG                   &0.5800                              &0.5886                              &\textbf{0.6394}                              &0.5810                                                                       \\
                        
  \midrule
\multirow{3}{*}{Llama3}      & CoT                   &\textbf{0.6386}                              &0.6386                              &0.6386                              &\textbf{0.6386}                                                                       \\
                             & MedRAG                &0.5127                              &0.5133                              &0.6170                              &0.5206                                                                       \\
                             & MoG                   &0.6221                              &\textbf{0.6415}                              &\textbf{0.6394}                              &0.6328                                                                       \\
                       
   \midrule
\multirow{3}{*}{Qwen1.5}     & CoT                   &0.3845                              &0.3845                              &0.3845                              &0.3845                                                                       \\
                             & MedRAG                &0.4399                              &\textbf{0.4623}                              &0.4499                              &0.4391                                                                       \\
                             & MoG                   &\textbf{0.6129}                              &0.4622                              &\textbf{0.5145}                              &\textbf{0.4547}                                                                       \\
                         
   \bottomrule
\end{tabular}
  \caption{Accuracy of Medical Question-Answering task with MoG (trained with different corpora)}
  \label{tab:mog_result_table_single_corpus}
\end{table*}

\section{Analysis on Samples Improved or Degraded}
\label{sec:analysis_sample}
\begin{figure*}[h!]
  \centering
  \includegraphics[width=\linewidth]{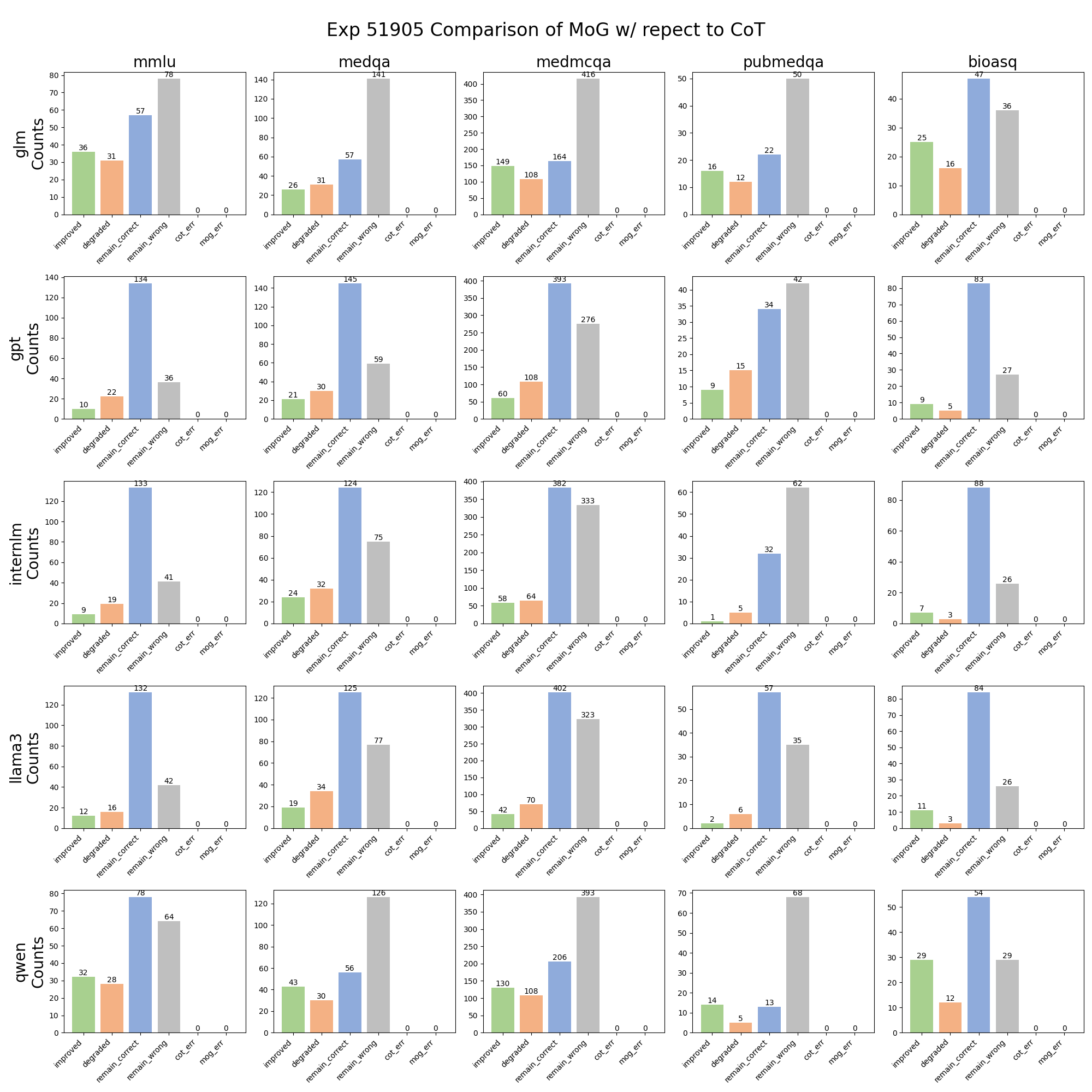}
  \caption{Number of samples improved or degraded after application of MoG}
  \label{fig:flag_plot}
\end{figure*}

We counted the number of samples of four categories before and after applying MoG. The four categories are 1. improved (CoT gives wrong answer, MoG gives correct answer); 2. degraded (CoT gives correct answer, MoG gives wrong answer); 3. remain\_correct (both CoT and MoG give correct answers); 4. remain\_wrong (both CoT and MoG give wrong answers). The results are visualized as Figure \ref{fig:flag_plot}. In this presented figure, the MoG is trained on the Wikipedia corpus. The pattern observed from this figure is similar to the ones shown in other test results. The zeroes in bars named ``cot\_err'' and ``mog\_err'' indicate that all the responses are correctly parsed when counting.

Around 10\% of the degraded samples are randomly chosen and manually verified. We confirmed that in most (95\%) degraded cases being verified, all the following statements are true: 
\begin{itemize}
\item several candidate snippets are retrieved correctly; 

\item top-2 candidate snippets are correctly selected with the weights calculated by the router; 

\item the prompt is correctly augmented and passed to backbone LLM; 

\item the LLM generates the response without error; 

\item the final choice (A/B/C/D or Yes/No) was correctly parsed. 
\end{itemize}
In other words, the LLM changed its answer based on the snippet retrieved, it is infected by the introduced noises. The degradation is caused by the fact that the tested RAG system lacks a noise filtering mechanism, rather than by the default of MoG or MoGG.

\section{Performance of MoGG on Medical QA Task (Other results)}
\label{sec:mogg_statpearls}
In this section, we present the experiment result of MoGG on the Medical QA task with StatPearls as the training corpus for the router. The results are grouped in Table \ref{tab:mogg_result_table_statpearls}.

\begin{table*}[ht]
  \centering
  \small
  \begin{tabular}{C{1cm}C{1cm}C{1.4cm}C{1.4cm}C{1.4cm}C{1.4cm}C{1.4cm}C{0.8cm}}
  \toprule
\multirow{2}{*}{LLM}       & \multirow{2}{*}{Method} & \multicolumn{6}{c}{MIRAGE Benchmark Dataset (Acc.)} \\

\cmidrule(r){3-8}
                           &                        & \small{MMLU}  & \small{MedQA} & \small{MedMCQA}   & \small{PubMedQA}  & \small{BioASQ}    & \small{Avg.} \\
   \midrule
\multirow{3}{*}{GLM3}      & CoT                     &\textbf{0.4901}\textcolor{g}{$\pm$0.04}&\textbf{0.3294}\textcolor{g}{$\pm$0.03} &\textbf{0.3799}\textcolor{g}{$\pm$0.02} &0.3900\textcolor{g}{$\pm$0.05}          &0.5645\textcolor{g}{$\pm$0.04}          &0.4181      \\
                           & MedRAG                  &\textbf{0.4901}\textcolor{g}{$\pm$0.04}&\textbf{0.3294}\textcolor{g}{$\pm$0.03} &\textbf{0.3799}\textcolor{g}{$\pm$0.02} &0.4400\textcolor{g}{$\pm$0.05}          &0.5323\textcolor{g}{$\pm$0.04}          &0.4343      \\
                           & MoG                     &0.4851\textcolor{g}{$\pm$0.04}         &0.2902\textcolor{g}{$\pm$0.03}          &\textbf{0.3799}\textcolor{g}{$\pm$0.02} &0.4400\textcolor{g}{$\pm$0.05}          &\textbf{0.6371}\textcolor{g}{$\pm$0.04} &\textbf{0.4465}      \\
                           & MoGG                    &0.4802\textcolor{g}{$\pm$0.04}         &0.3059\textcolor{g}{$\pm$0.03}          &0.3668\textcolor{g}{$\pm$0.02}          &\textbf{0.4600}\textcolor{g}{$\pm$0.05} &0.5806\textcolor{g}{$\pm$0.04}          &0.4387      \\       
   \midrule
\multirow{3}{*}{GPT-3.5}   & CoT                     &\textbf{0.7624}\textcolor{g}{$\pm$0.03}&\textbf{0.6706}\textcolor{g}{$\pm$0.03} &\textbf{0.5866}\textcolor{g}{$\pm$0.02} &\textbf{0.4200}\textcolor{g}{$\pm$0.05} &\textbf{0.7339}\textcolor{g}{$\pm$0.04} &\textbf{0.6347}      \\
                           & MedRAG                  &0.6881\textcolor{g}{$\pm$0.03}         &0.6549\textcolor{g}{$\pm$0.03}          &0.4552\textcolor{g}{$\pm$0.02}          &0.2600\textcolor{g}{$\pm$0.04}          &0.5565\textcolor{g}{$\pm$0.04}          &0.5229      \\
                           & MoG                     &0.7277\textcolor{g}{$\pm$0.03}         &0.6471\textcolor{g}{$\pm$0.03}          &0.5400\textcolor{g}{$\pm$0.02}          &0.3200\textcolor{g}{$\pm$0.05}          &0.7258\textcolor{g}{$\pm$0.04}          &0.5921      \\
                           & MoGG                    &\textbf{0.7624}\textcolor{g}{$\pm$0.03}&0.6314\textcolor{g}{$\pm$0.03}          &0.5460\textcolor{g}{$\pm$0.02}          &0.3700\textcolor{g}{$\pm$0.05}          &0.7177\textcolor{g}{$\pm$0.04}          &0.6055      \\
   \midrule
\multirow{3}{*}{InternLM}  & CoT                     &\textbf{0.7228}\textcolor{g}{$\pm$0.03}&\textbf{0.6000}\textcolor{g}{$\pm$0.03} &0.5352\textcolor{g}{$\pm$0.02}          &0.3700\textcolor{g}{$\pm$0.05}          &0.7339\textcolor{g}{$\pm$0.04}          &0.5924      \\
                           & MedRAG                  &0.6584\textcolor{g}{$\pm$0.03}         &0.5137\textcolor{g}{$\pm$0.03}          &0.3596\textcolor{g}{$\pm$0.02}          &0.1500\textcolor{g}{$\pm$0.04}          &0.5081\textcolor{g}{$\pm$0.04}          &0.4380      \\
                           & MoG                     &0.6881\textcolor{g}{$\pm$0.03}         &0.5961\textcolor{g}{$\pm$0.03}          &\textbf{0.5448}\textcolor{g}{$\pm$0.02} &0.3800\textcolor{g}{$\pm$0.05}          &\textbf{0.7339}\textcolor{g}{$\pm$0.04} &0.5886      \\
                           & MoGG                    &0.6782\textcolor{g}{$\pm$0.03}         &0.5725\textcolor{g}{$\pm$0.03}          &0.3500\textcolor{g}{$\pm$0.02}          &\textbf{0.7500}\textcolor{g}{$\pm$0.05} &\textbf{0.7500}\textcolor{g}{$\pm$0.04} &\textbf{0.6201}      \\
  \midrule
\multirow{3}{*}{Llama3}    & CoT                     &0.7277\textcolor{g}{$\pm$0.03}         &\textbf{0.6392}\textcolor{g}{$\pm$0.03} &\textbf{0.5663}\textcolor{g}{$\pm$0.02} &0.5900\textcolor{g}{$\pm$0.05}          &0.7177\textcolor{g}{$\pm$0.04}          &\textbf{0.6482}      \\
                           & MedRAG                  &0.6089\textcolor{g}{$\pm$0.03}         &0.5882\textcolor{g}{$\pm$0.03}          &0.4170\textcolor{g}{$\pm$0.02}          &0.3800\textcolor{g}{$\pm$0.05}          &0.5242\textcolor{g}{$\pm$0.04}          &0.5133      \\
                           & MoG                     &\textbf{0.7376}\textcolor{g}{$\pm$0.03}&0.5961\textcolor{g}{$\pm$0.03}          &0.5317\textcolor{g}{$\pm$0.02}          &\textbf{0.6000}\textcolor{g}{$\pm$0.05} &\textbf{0.7419}\textcolor{g}{$\pm$0.04} &0.6415      \\
                           & MoGG                    &0.7030\textcolor{g}{$\pm$0.03}         &0.6275\textcolor{g}{$\pm$0.03}          &0.5436\textcolor{g}{$\pm$0.02}          &0.5900\textcolor{g}{$\pm$0.05}          &0.7258\textcolor{g}{$\pm$0.04}          &0.6380      \\
   \midrule
\multirow{3}{*}{Qwen1.5}   & CoT                     &0.4604\textcolor{g}{$\pm$0.04}         &0.3255\textcolor{g}{$\pm$0.03}          &0.3883\textcolor{g}{$\pm$0.02}          &0.2000\textcolor{g}{$\pm$0.04}          &0.5484\textcolor{g}{$\pm$0.04}          &0.3845      \\
                           & MedRAG                  &\textbf{0.5495}\textcolor{g}{$\pm$0.04}&\textbf{0.4471}\textcolor{g}{$\pm$0.03} &0.4146\textcolor{g}{$\pm$0.02}          &\textbf{0.3600}\textcolor{g}{$\pm$0.05} &0.5403\textcolor{g}{$\pm$0.04}          &0.4623      \\
                           & MoG                     &0.5446\textcolor{g}{$\pm$0.04}         &0.3882\textcolor{g}{$\pm$0.03}          &\textbf{0.4337}\textcolor{g}{$\pm$0.02} &0.3400\textcolor{g}{$\pm$0.05}          &0.6048\textcolor{g}{$\pm$0.04}          &0.4623      \\
                           & MoGG                    &0.5446\textcolor{g}{$\pm$0.04}         &0.4157\textcolor{g}{$\pm$0.03}          &0.4253\textcolor{g}{$\pm$0.02}          &0.3500\textcolor{g}{$\pm$0.05}          &\textbf{0.6290}\textcolor{g}{$\pm$0.04} &\textbf{0.4729}      \\
   \bottomrule
\end{tabular}
  \caption{Accuracy of Medical Question-Answering task with MoGG (trained with StatPearls), best results marked in \textbf{bold}}
  \label{tab:mogg_result_table_statpearls}
\end{table*}

\section{MoGG's Abaltion test}
\label{sec:mogg_ablation}
In this section, we rerun the MoGG without the router on Llama 3 to highlight its effectiveness. The results of MoGG's ablation test is in Table \ref{tab:llm_performance} (highlighted in bold). The results are similar to MedRAG, because, without the router, MoGG is essentially the same as MedRAG, the added flexibility of the graph will only show when used with a router.

\begin{table*}[ht]
    \centering
    \begin{tabular}{@{}lcccccc@{}}
        \toprule
        Method & MMLU & MedQA & MedMCQA & PubMedQA & BioASQ & Avg. \\ \midrule
                CoT & 0.7079 & 0.6431 & 0.5663 & 0.5500 & 0.7258 & 0.6386 \\
                MedRAG & 0.6485 & 0.5961 & 0.4146 & 0.3800 & 0.5242 & 0.5127 \\
                MoG & 0.7228 & 0.6196 & 0.5484 & 0.5100 & 0.7097 & 0.6221 \\
                \textbf{MoGG w/o router} & \textbf{0.6683} & \textbf{0.5451} & \textbf{0.4182} & \textbf{0.3800} & \textbf{0.5484} & \textbf{0.5120} \\
                MoGG & 0.7070 & 0.5961 & 0.5460 & 0.5200 & 0.7661 & 0.6262 \\ \bottomrule
    \end{tabular}
    \caption{Ablation test with Llama3 as backbone LLM}
    \label{tab:llm_performance}
\end{table*}

\section{MoG Trained With Only Textbooks Corpus}
\label{sec:mog_only_textbooks}
In this Appendix, we present the results of training MoG only with Textbooks corpus. As Textbooks are only a subset of MedCorp, the performance is not as good as the ones shown in Table \ref{tab:mog_result_table}. Compared with Table \ref{tab:mog_only_textbooks} and Table \ref{tab:mogg_result_table_textbooks}, we can see the advantage of MoGG more clearly.

\begin{table*}[ht]
    \centering
    \small
    \begin{tabular}{C{1cm}C{1cm}C{1.4cm}C{1.4cm}C{1.4cm}C{1.4cm}C{1.4cm}C{0.8cm}}
        \toprule
        \multirow{2}{*}{LLM}       & \multirow{2}{*}{Method} & \multicolumn{6}{c}{MIRAGE Benchmark Dataset (Acc.)} \\
        \cmidrule(r){3-8}
                                   &                        & \small{MMLU}  & \small{MedQA} & \small{MedMCQA}   & \small{PubMedQA}  & \small{BioASQ}    & \small{Avg.} \\
        \midrule
        \multirow{3}{*}{GLM3}  & MedRAG                  & 0.4802         & 0.3569       & 0.3811            & 0.3600           & 0.5565           & 0.4269      \\
                                   & MoG                     & 0.5545         & 0.2941       & 0.3548            & 0.4700           & 0.5726           & 0.4492      \\
        \midrule
        \multirow{3}{*}{GPT-3.5}   & MedRAG                  & 0.7277         & 0.6745       & 0.4468            & 0.2600           & 0.5161           & 0.5250      \\
                                   & MoG                     & 0.7525         & 0.6667       & 0.5603            & 0.5200           & 0.7016           & 0.6122      \\
        \midrule
        \multirow{3}{*}{InternLM}  & MedRAG                  & 0.6188         & 0.5569       & 0.3118            & 0.1400           & 0.4839           & 0.4223      \\
                                   & MoG                     & 0.7277         & 0.5725       & 0.5281            & 0.3400           & 0.7339           & 0.5804      \\
        \midrule
        \multirow{3}{*}{Llama3}    & MedRAG                  & 0.6485         & 0.5961       & 0.4146            & 0.3800           & 0.5242           & 0.5127      \\
                                   & MoG                     & 0.7228         & 0.6196       & 0.5484            & 0.5100           & 0.7097           & 0.6221      \\
        \midrule
        \multirow{3}{*}{Qwen1.5}   & MedRAG                  & 0.5941         & 0.4000       & 0.3835            & 0.3300           & 0.4919           & 0.4399      \\
                                   & MoG                     & 0.5792         & 0.3843       & 0.4110            & 0.3300           & 0.6129           & 0.4635      \\
        \bottomrule
    \end{tabular}
    \caption{Accuracy of Medical Question-Answering task with MoG (trained with only Textbooks)}
    \label{tab:mog_only_textbooks}
\end{table*}

\section{Detailed Analysis of Time Consumption}
\label{sec:time_consumption}

In this appendix, we present the detailed experiment results of execution time. We measure the wall-clock execution time with different number of granularities using Llama3 as the backbone LLM. The results organized in Table \ref{tab:time_consumption} show that:

(1) the introduction of the router module will effectively increase the inference time by about 60\%; 

(2) increasing the number of granularity levels will only increase a marginal increase in execution time. The reason is that the bottleneck of execution time is the API calling time of backbone LLMs.

\begin{table*}[ht]
    \centering
    \begin{tabular}{@{}lcccccc@{}}
        \toprule
        \# (Granularity) & MMLU & MedQA & MedMCQA & PubMedQA & BioASQ & Global Avg. \\ \midrule
        1 w/o router & 6.61 & 9.02 & 10.01 & 7.40 & 7.53 & 8.12 \\
        1 w/ router  & 10.15 & 10.76 & 15.81 & 15.99 & 13.09 & 13.16 \\
        2            & 13.25 & 12.60 & 20.53 & 17.30 & 14.98 & 15.73 \\
        3            & 16.30 & 16.01 & 19.03 & 16.73 & 17.12 & 17.04 \\
        4            & 13.43 & 20.65 & 22.72 & 18.93 & 14.05 & 17.96 \\
        5            & 13.20 & 20.43 & 23.00 & 19.36 & 13.66 & 17.93 \\ \bottomrule
    \end{tabular}
    \caption{Performance Metrics by Granularity}
    \label{tab:time_consumption}
\end{table*}

\end{document}